\documentclass[acmlarge]{acmart}

\usepackage{appendix}
\usepackage[utf8]{inputenc} 
\usepackage{hyperref}       
\usepackage{url}            
\usepackage{amsfonts}       
\usepackage{nicefrac}       
\usepackage{microtype}      
\usepackage{graphicx}
\usepackage{natbib}
\usepackage{doi}
\usepackage{float}
\usepackage{tablefootnote}

\makeatletter
\newcommand*\breveunder[1]{\oalign{#1\crcr\hidewidth\ltx@sh@ft{-3ex}%
   \vbox to .2ex{\hbox{\u{}}\vss}\hidewidth}}
\makeatother

\title{DeepScribe: Localization and Classification of Elamite Cuneiform Signs Via Deep Learning}

\acmJournal{JOCCH}

 \setcopyright{none}

 \received{22 December 2022}

\author{Edward C. Williams}
\email{eddiecwilliams@gmail.com}
\orcid{0000-0002-5812-2831}
\affiliation{%
  \institution{Independent Researcher}
  \city{Los Angeles}
  \state{California}
  \country{USA}
}
\author{Grace Su}
\email{g.su@columbia.edu}
\orcid{0000-0001-5102-9258}
\affiliation{%
  \institution{Department of Computer Science, Columbia University}
  \city{New York City}
  \state{New York}
  \country{USA}
}
\author{Sandra R. Schloen}
\email{sschloen@uchicago.edu}
\orcid{0000-0002-0695-6797}
\affiliation{%
  \institution{Forum for Digital Culture, University of Chicago}
  \city{Chicago}
  \state{Illinois}
  \country{USA}
}
\author{Miller C. Prosser}
\email{m-prosser@uchicago.edu}
\orcid{0000-0002-6344-080X}
\affiliation{%
  \institution{Forum for Digital Culture, University of Chicago}
  \city{Chicago}
  \state{Illinois}
  \country{USA}
}
\author{Susanne Paulus}
\email{paulus@uchicago.edu}
\orcid{1234-5678-9012}
\affiliation{%
  \institution{Institute for the Study of Ancient Cultures, University of Chicago}
  \city{Chicago}
  \state{Illinois}
  \country{USA}
}
\author{Sanjay Krishnan}
\email{skr@uchicago.edu}
\orcid{0000-0001-6968-4090}
\affiliation{%
  \institution{Department of Computer Science, University of Chicago}
  \city{Chicago}
  \state{Illinois}
  \country{USA}
}

\begin{document}
\begin{abstract}
Twenty-five hundred years ago, the ``paperwork'' of the Achaemenid Empire was recorded on clay tablets. In 1933, archaeologists from the University of Chicago’s Institute for the Study of Ancient Cultures (ISAC, formerly Oriental Institute) found tens of thousands of these tablets and fragments during the excavation of Persepolis. Many of these tablets have been painstakingly photographed and annotated by expert cuneiformists, and now provide a rich dataset consisting of over 5,000 annotated tablet images and 100,000 cuneiform sign bounding boxes encoding the Elamite language. We leverage this dataset to develop DeepScribe, the first computer vision pipeline capable of localizing Elamite cuneiform signs and providing suggestions for the identity of each sign. We investigate the difficulty of learning subtasks relevant to Elamite cuneiform tablet transcription on ground-truth data, finding that a RetinaNet object detector achieves a localization mAP of 0.78 and a ResNet classifier achieves a top-5 sign classification accuracy of 0.89. The end-to-end pipeline achieves a top-5 classification accuracy of 0.80. As part of the classification module, DeepScribe groups cuneiform signs into morphological clusters. We consider how this automatic clustering approach differs from the organization of standard, printed sign lists and what we learn from it. These components, trained individually, are sufficient to produce a system that can analyze photos of cuneiform tablets from the Achaemenid period and provide useful transliteration suggestions to researchers. We evaluate the model’s end-to-end performance on locating and classifying signs, providing a roadmap to a linguistically-aware transliteration system, then consider the model's potential utility when applied to other periods of cuneiform writing. 

\end{abstract}
\maketitle
\section{Introduction}

\begin{figure}
\centering
\includegraphics[width=1.0\textwidth]{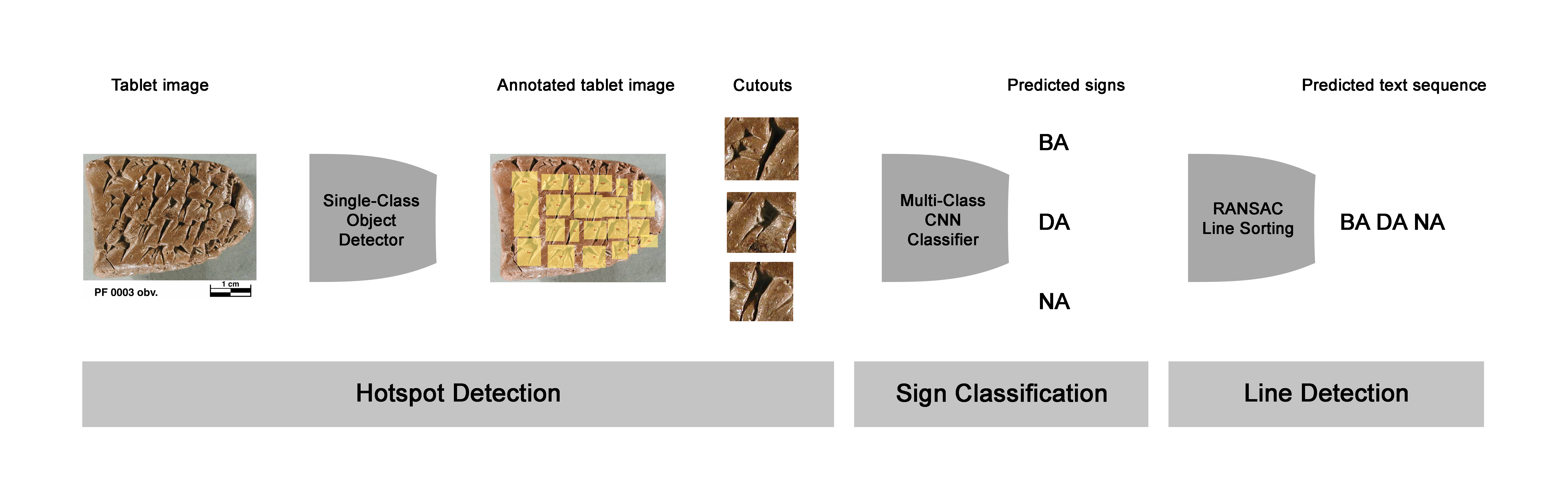}
\caption{DeepScribe Vision Pipeline}
\label{fig:pipeline}
\end{figure}

Written documents are a central pillar of the study of ancient history. 
Scholars rely on these primary sources to provide insight into social, political, economic and cultural history. 
For the ancient Near East and Mediterranean, our historical sources are often those inscriptions that have survived on durable materials like clay. Cuneiform texts contain economic records, royal decrees, personal letters, epic poems and stories, among many other types of communication and record-keeping spanning thousands of years of history.
This paper considers one corpus of written records of the Achaemenid Empire: a large archive of administrative texts from Persepolis \cite{Hallock_1969} \cite{stolper_report} \cite{Stolper_2007} \cite{henkelman2013administrative}. The Persepolis Fortification Archive (PFA) is primarily written in Elamite, a language isolate written and spoken in Iran in antiquity. Other texts are written using in Aramaic, Old Persian, Greek, Akkadian, Phrygian, and Demotic, but
it is the primary corpus of cuneiform Elamite texts that serves as the core of this study.  Over 5000 additional Elamite tablets are present with no or partial annotation (transliteration but no bounding boxes) \footnote{As of October 2023.}. We leverage the efforts of the PFA editorial team -  which has produced thousands of digital images of tablets annotated with sign-level bounding boxes (referred to as ``hotspots'') - to assist in the transliteration of other tablets via machine learning. Hundreds of thousands more texts from other periods and languages are thought to exist \cite{altorientalistik2010umfang}, possibly numbering in the millions \cite{reade2017manufacture}. We focus specifically on the problem of transliterating Elamite text, although similarities between Elamite cuneiform and texts from other periods and languages do exist.


The process of transcribing a cuneiform text is slow and laborious, requiring a great deal of training and expertise. Producing a sign-by-sign transliteration of a text could easily take an experienced scholar multiple days. 
Many researchers wish also to annotate digital images to indicate the location of a sign on a tablet, whether to use as a pedagogical tool or to more clearly communicate their interpretation. 
Even a dedicated student worker would find it hard to annotate images of more than a few tablets per day.
Enabling the rapid search of these corpora and the indexing of new cuneiform text corpora via automated annotation would speed up the research process and provide a great boon to historians of the ancient Near East. Our purpose is not to remove the skilled cuneiformist from the process --- but to spare them the most tedious initial work.  

Current machine learning methods are able to recognize modern handwritten text with accuracy and performance close to that of humans, due to developments in deep neural networks and large-scale datasets with a variety of data instances \cite{DBLP:journals/corr/abs-1910-00663}. Character classification datasets such as MNIST \cite{lecun-mnisthandwrittendigit-2010} possess tens of thousands of carefully curated training examples, while in-context handwriting recognition datasets such as IAM \cite{Marti2002TheIA} contain over 1500 scanned documents with line-by-line annotations. However, using a model to automatically transcribe an ancient text is substantially more challenging than typical problems in handwriting recognition. First, the fragmentary condition of these clay tablets leads to inherent data quality problems where even a trained human may not be able to determine what sign a group of wedges represents. Second, handwriting recognition methods are normally applied to two-dimensional printed or handwritten text on flat surfaces. Because cuneiform signs are produced by impressing a reed stylus into wet clay, the signs are inherently three-dimensional. It is important to note that the tablet surface is also a variable three-dimensional object, making digital photography challenging. A cuneiform text is most easily read when the tablet is held in the hand and rotated dynamically to produce variable light and shadow across the tablet surface. 
Finally, there is simply a lack of comparable large-scale datasets for ancient Near Eastern scripts. Data archives for these scripts exist, such as ORACC \cite{oracc}, which contains thousands of transliterations,  and CDLI \cite{cdli}, which contains  thousands of cuneiform tablet images, some paired with transliteration. However, these archives do not possess sign-level bounding box annotations that enable the straightforward use of modern object detection methods. In addition, these corpora primarily contain text written in Akkadian or Sumerian, giving them limited utility for Elamite transliteration.

In this paper, we present a modular computer vision pipeline that leverages the richly annotated PFA dataset to localize and classify Elamite cuneiform signs. We use state-of-the-art computer vision methods, which can be applied largely off-the-shelf due to the quality and type of annotations in the PFA. This is the first work of its kind that exploits the potential of this dataset, as well as the first to address the automated recognition of Elamite cuneiform. 
We then propose a multi-stage computer vision pipeline using a RetinaNet \cite{DBLP:journals/corr/abs-1708-02002} object detector to identify rectangular image regions containing a single cuneiform sign, then feed identified signs to a 141-class image classifier trained on ground truth sign data. Finally, we convert hotspot locations to discrete lines using a variant of the Sequential RANSAC algorithm \cite{938625}. 

We perform a detailed analysis of each stage in this pipeline, with the goal of understanding the difficulty of each subtask (sign detection, sign classification, sequence transliteration) to determine the practical utility of such a pipeline. We find that the computer vision methods are able to localize signs and provide ranked suggestions as to the identity of each sign. We anticipate that these vision-based methodologies will provide a useful ``initial guess" for scholars annotating new Elamite tablets.

Additionally, we evaluate the end-to-end sequence transliteration performance of this vision-based method. While the hotspot detector and sign classifier in concert are able to provide useful suggestions to cuneiformists, a fully end-to-end transliteration system will likely require explicit modeling of linguistic context. We outline a path forward to incorporate linguistic information in future work to eventually produce complete tablet transliterations from an unannotated image. We also provide some preliminary evidence that sign detection models trained on Elamite cuneiform have some applicability to other periods and languages, although we leave further exploration of cross-language cuneiform model transfer to future work.

\section{Background and Related Work}
\label{sec:related}

Below we provide a summary of related work in the domain of text recognition broadly, and the specific application of computer vision methods to cuneiform data. We note that due to the variety of languages encoded by various cuneiform scripts, specific quantitative results from cuneiform datasets containing different languages are not directly comparable. For example, related work on English or Sumerian text provides an intuitive sense for results on a ``clear" or ``challenging" corpus, but not an exact benchmark that can be applied to Elamite text.


\subsection{Text Recognition}
Tools to partially automate or aid human annotation would enable the rapid creation of richly annotated digital archives. The success of machine learning algorithms in image analysis tasks both inside and outside of document recognition \cite{alex1imagenet} \cite{Zhou_2021} \cite{DBLP:journals/corr/abs-1708-02002} \cite{DBLP:journals/corr/RenHG015} \cite{DBLP:journals/corr/abs-1910-00663} \cite{hodel_schoch_schneider_purcell_2021} and the proliferation of open-source software tools for developing machine learning products \cite{sklearn_api}\cite{wu2019detectron2}\cite{NEURIPS2019_9015} present the possibility of using these tools for the analysis of historical documents. 

Extensive research on computer vision methods for document recognition has been conducted over the past 30 years, resulting in software packages that are widely used in both research and industry \cite{smith2007overview} \cite{lee2012improving}. Such methods rely on software pipelines that identify lines of text, segment lines into words or characters, and identify characters. Handwritten text recognition, due to the relative irregularity of handwritten characters, typically requires dedicated processing pipelines \cite{DBLP:journals/corr/abs-1910-00663} \cite{renton2018fully} \cite{https://doi.org/10.48550/arxiv.0705.2011} \cite{hodel_schoch_schneider_purcell_2021} but can be performed effectively by an automated system, with character error rates under 8\% \cite{DBLP:journals/corr/Bluche16}. Full-page handwriting recognition methods typically segment an image into text lines via object detection methods that are then decoded using an alignment-based loss function \cite{Wigington2018StartFR} \cite{DBLP:journals/corr/abs-1910-00663}, although segmentation-free methods exist and are known to be effective \cite{DBLP:journals/corr/Bluche16}. We apply a similar factorization, first using an object detector to identify cuneiform signs and then to classify them, although our access to per-character bounding boxes allows us to avoid the problem of line segmentation. In the context of the PFA, text lines are often curved, which presents a problem for standard line segmentation techniques. Methods for detecting curved text lines exist \cite{DBLP:journals/corr/abs-1903-09837}, but the PFA dataset we utilize does not possess line-level annotations. 
Training modularized subcomponents of OCR systems has precedent in the transliteration of modern texts \cite{DBLP:journals/corr/abs-1910-00663}, but the primary purpose it serves here is to improve our understanding of the novel PFA dataset from the perspective of computer vision. Modeling OCR as a combination of object recognition and transliteration tasks is fairly common in the field \cite{Wigington2018StartFR} \cite{DBLP:journals/corr/abs-1910-00663}, but it is relatively rare to perform OCR on the word or character level for Latin scripts. However, prior work transliterating Chinese characters \cite{https://doi.org/10.48550/arxiv.1803.00085} has adopted a per-character object recognition approach, showing that different linguistic contexts often require customized pipelines. 

The success of these methods has inspired their application to historical document analysis, developing tools to automatically extract useful semantic information from documents that are typically only readable by specialized scholars. For instance, \cite{9041761} describes a  text recognition system intended for use on handwritten historical documents, as well as a publicly available software tool that allows for fine-tuning on user-provided corpora. 

\subsection{Cuneiform Text Recognition}

\cite{10.1145/3491239} provides an excellent review of projects applying techniques from computer vision to cuneiform tablet images, as well as a typology of the diverse methodologies that have been adopted to analyze these data sets. While the relatively small datasets of 3D-scanned or photographed cuneiform script typically limit the utility of the large deep learning models that are often successful in addressing computer vision problems, digital humanists and computer scientists have spent decades building automated systems that can leverage what data exists in archives such as the Cuneiform Digital Library Initiative and ORACC.  \cite{bogacz2018} uses a keypoint-based method to identify cuneiform signs on vector graphics representations of cuneiform text, but this requires the manual creation of a vectorized transcript before any analysis can be performed. 

The recent work of \cite{dencker2020deep} uses a weakly-supervised  algorithm to learn to detect cuneiform signs from a 2D image dataset of Neo-Assyrian tablets found on ORACC \cite{oracc}, which are linguistically distinct from our Elamite tablet set. Due to the lack of annotated sign localizations in the ORACC dataset, they develop an iterative learning procedure using a Single-Shot Detector (SSD) object detector algorithm \cite{Liu_2016} along with line alignment techniques to iteratively propose, filter, and retrain on automatically generated sign localizations. Additionally, this work includes a Conditional Random Field (CRF) \cite{laffertycrf2001} sequence-modeling step as the final stage of the annotation-alignment pipeline, but this uses geometric and spatial information to match predicted and classified regions to transliteration characters, rather than using contextual information to refine the raw predictions of a sign detector. They also find a boost in accuracy when their models are provided with a small set (< 60) of fully annotated tablet images. Our approach distinguishes itself from this effort firstly by utilizing a novel corpus of Elamite tablets, and secondly by directly training on localized sign annotations rather than leveraging weak supervision. However, we see several promising avenues for future works in the combination of these two approaches, such as the leveraging of a model trained on a large dataset of annotated signs to initialize weakly-supervised learning methods on a low-resource dataset, although there are many questions to be answered on the efficacy of cross-linguistic transfer. Given the large linguistic differences in the two corpora (Neo-Assyrian Akkadian and Elamite), metrics are not directly comparable between the two methods.  

An orthogonal approach that that may be better suited to text written on curved tablets uses 3D scanning information as input to cuneiform tablet analysis. However, acquiring 3D-scanned data requires far more time, labor, and cost than 2D image data. In addition, acquiring permission to 3D scan tablets from museums is often difficult.  \cite{bogacz2015character} uses 3D models of cuneiform tablets to extract vector drawings and retrieve the corresponding cuneiform characters. \cite{bogacz2020period} also uses 3D data, but predicts the time period of a cuneiform tablet instead by training a Convolutional Neural Network (CNN) on a partially transliterated cuneiform tablet 3D surface mesh dataset. \cite{kriege2018graph} addresses the geometric nature of cuneiform script by classifying signs based on graph model data of cuneiform signs extracted from 3D scans using the method of \cite{fisseler2013towards}, but still must rely on a small set of annotated data. \cite{rusakov2019generating} attempts to resolve this by generating artificial 2D projections of 3D data using a Generative Adversarial Network (GAN) \cite{goodfellow2014generative}, although annotations still must be provided manually for downstream classification tasks. We instead focus on the problem of recognizing 2D signs from flat images of a tablet, due to the nature of the annotations present in our dataset. 

\section{Data}

In this section, we describe the dataset, the preprocessing steps applied use the dataset in our computer vision pipeline, and the metrics and evaluation scheme used to measure the performance of the pipeline.

\subsection{The Persepolis Fortification Archive}

In 1933, the University of Chicago’s Institute for the Study of Ancient Cultures (ISAC, formerly Oriental Institute) discovered a large cache of cuneiform tablets at the site of Persepolis. The tablets were discovered in two rooms inside a fortification wall and were dubbed the Persepolis Fortification Archive (PFA) \cite{Stolper_2007}. Scholars at the ISAC and world-wide have been studying and publishing editions of these tablets since 1937 \cite{Hallock_1969}. The tablets provide valuable insight into the administration and economic history of the Achaemenid Empire, and their digitization enables the rapid search and indexing of this large corpus. The majority of the PFA texts record the distribution of commodities and allocation of resources, which has led to a greater understanding of the internal administrative details structure of Persia \cite{henkelman2013administrative}. In 2006, a renewed effort to document and publish the texts led to the creation of the PFA dataset as it exists today. University of Chicago professor Matthew W. Stolper assembled an international editorial and technology team to work on the PFA project. The team included specialists in Elamite and Aramaic language, seal iconography, digital imaging, and data representation. The editorial team worked on new editions of the texts. The seal iconography team faced the massive task of identifying, classifying, and studying the thousands of unique seal impressions recorded on the tablets \cite{Solr-4582863}. The photography team instantiated three separate digital photography modalities: conventional digital photography, high-resolution scanning under filtered light, and Polynomial Texture Mapping (also known as Reflectance Transformation Imaging) \cite{stolper_report}. The database team consulted with the rest of the PFA team to record all of this information in an integrated system called the OCHRE database platform.

\begin{figure}
\centering
\includegraphics[width=0.5\textwidth]{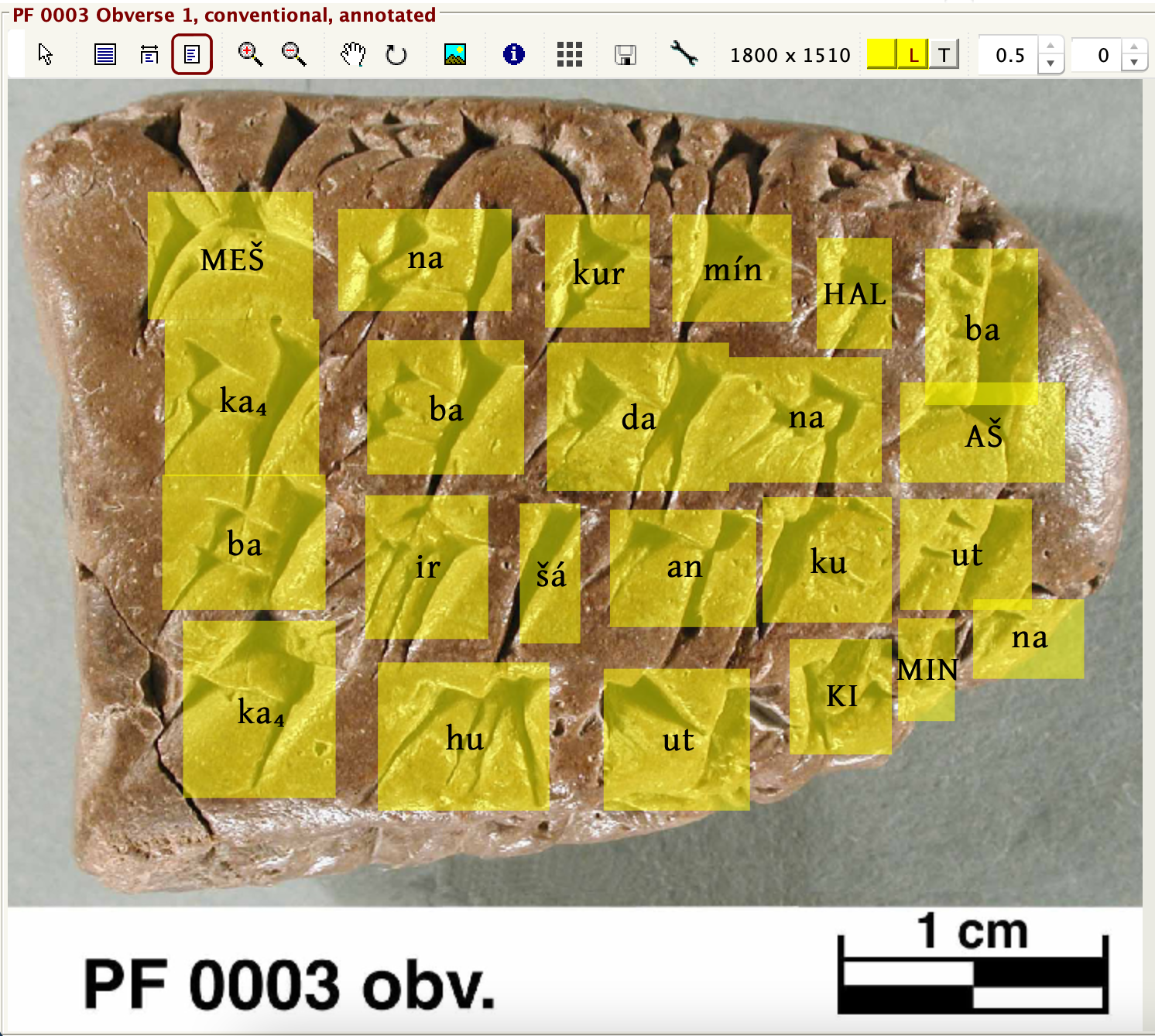}
\caption{PFA Tablet Image with Annotated Hotspots viewed in OCHRE}
\label{fig:pfa_tablet}
\end{figure}

\subsection{Dataset Production}

The PFA dataset was annotated with character bounding-box level annotations using the Online Cultural and Historical Research Environment (OCHRE, \url{ochre.uchicago.edu}), a research database platform designed for humanities and social science data. Each clay tablet, each digital photograph, and even each cuneiform sign in each text is a discrete database object, described by metadata and associated with hundreds or thousands of other database items. 


In OCHRE, the PFA editorial team has created text editions in which each cuneiform sign was recorded as a discrete database item. Each sign reading represents an interpretation by a PFA editor, reached through careful examination of the clay tablet under magnification which leads to a reading of the signs and translation. As shown in Figure \ref{fig:pfa_tablet}, PFA team members use OCHRE to associate manually-created tablet image hotspots with transliterations of the text in question. Linguistic information is therefore incorporated at the sign annotation level even if individual signs may be difficult for humans to read in isolation - i.e., if a sign is difficult to read but its value is inferred via linguistic context, its bounding box will still be present.
To account for the polyvalent nature of the logographic and syllabic cuneiform writing system, each sign is linked with both its sign identity and its syllabic value. For example, the sign GIŠ can communicate syllabic values \textit{is/ez}, \textit{giš} and \textit{bil} as well as the logographic usage meaning "wood" and the determinative usage which refers to wood and wood-based products. The identification of the value of a sign must be determined by context. In this stage of the DeepScribe project, we set aside the sign–to–value mapping problem and focus entirely on sign identity prediction. In future work, we plan to take advantage of the extensive PFA glossary of Elamite words in OCHRE to suggest sign values and word identifications.
The original intent of the PFA annotation effort was to create a pedagogical tool to help teach the reading of Elamite. Instead of providing students with a hand-drawn version of a cuneiform text with highly regularized cuneiform signs—as it is often the case in the classroom—these annotated images would allow the student to engage with the cuneiform text as recorded by the scribe. 

The sign-bounding-box level annotations provide a dataset that is well suited for the application of off-the-shelf computer vision algorithms. Unlike other labeled cuneiform datasets which typically consist of whole-tablet images paired with transliterations \cite{oracc}, the PFA tablet images in OCHRE are labeled much like object recognition datasets such as COCO (Common Objects in Context) \cite{https://doi.org/10.48550/arxiv.1405.0312}. Each image of a PFA tablet is paired with a set of labeled bounding boxes, which can either be used out-of-context to classify cuneiform signs, or in-context for sign bounding box detection. We explore both in this work as part of a modular recognition pipeline. The dataset is composed of images of tablets themselves, not line drawings or artificial constructs - the dataset even  contains natural handwriting variation among scribes. This allows computational models to learn directly from cuneiform text as it was written by scribes. We anticipate that this will produce models capable of recognizing Elamite cuneiform text \textit{in situ}, without requiring human scholars to perform tedious pre-processing before annotations can be generated.

\subsection{Dataset Summary}
\label{sec:dataset_summary}
The full dataset used in this work consists of 5007 images of 1360 unique tablets, where each image contains a set of hotspot region annotations. We find that the relative frequencies of the 141 unique Elamite signs contained within the dataset are highly unbalanced, with the vast majority of signs being fairly poorly attested. As expected from a natural language dataset, the dataset displays an approximately Zipfian distribution \cite{piantadosi2014zipf}. High performance on high-frequency signs can easily “wash out” low performance on low-frequency signs on aggregate metrics. As would be expected from a power-law distributed data set, the top 50 most frequent signs cover 86.1\% of the dataset, indicating that the remaining majority of sign classes are poorly represented. As such, there may be a trade-off between a system focused on low-resource signs and a system capable of automatically annotating ``easy'' or high-resource signs then flagging others for human intervention. We explore the relationship between class frequency and performance in Section \ref{section:cls_perf}. We perform basic processing to consolidate redundant class labels and automatically remove unannotated image regions, described in more detail in Appendix \ref{sec:dset}.

\subsection{Public Distribution}
A public subset of the dataset, consisting of 1239 tablet images, is available in JSON format \footnote{See \url{https://pi.lib.uchicago.edu/1001/org/ochre/ad64d9db-de4a-426c-81d5-10bf292ffd0e} for links to the project code and data.}. The ``full" dataset used in this work contains unpublished tablets courtesy of the PFA project. These tablets can only be released after the primary publication of the cuneiform texts. We provide code examples of training and evaluating models on public data using Google Colaboratory \cite{google-colab}. We find that models trained only on the public subset achieve somewhat reduced performance on held-out test data, consistent with the scaling explored in Sections \ref{sec:det-analysis} and \ref{section:cls_perf}. 



\subsection{Metrics and Evaluation}
\label{sec:metrics}

We use a series of metrics designed to evaluate model performance on subtasks within our dataset, in addition to the general task of end-to-end cuneiform sign localization. 

\subsubsection{Sign Detection}
Before constructing any end-to-end system, we would like to know, primarily, how easily an object detector can localize \textit{any} cuneiform sign, regardless of identity, and how well an off-the-shelf classification algorithm can recognize cuneiform signs when provided with ``correct'' (i.e., human-generated) image regions. For the former task, we rely on standard object detection metrics, prioritizing the single-class Average Precision (AP) used to evaluate object detectors on other datasets \cite{DBLP:journals/corr/abs-1708-02002}. This metric is computed with respect to an Intersection over Union (IoU) threshold that demarcates whether a prediction is a true or false positive. Per-class AP values are computed, and then averaged over all classes (in the multi-class case). These values can be computed over varying IoU thresholds, but we focus on the AP@50 (50\% IoU overlap) metric. State of the art multi-class object detectors achieve AP@50 metrics upwards of 61\% on large datasets such as COCO \cite{DBLP:journals/corr/abs-1708-02002}.

\subsubsection{Sign Classification}

For sign classification, we focus on the top-k accuracy. In this setting, a prediction is marked as a true positive if the correct sign is in the top-k of a ranked list of predictions. For example, if a the top-ranked prediction is not correct, but the correct sign is highly ranked (i.e., the 4th highest ranked), then the prediction would be marked as "correct" from the perspective of the top-5 accuracy. This provides two key pieces of insight: firstly, understanding whether or not the network is approximately correct in cases where it incorrectly predicts the top-ranked sign, and secondly, determining how practically useful a ranked list of signs provided by the model would be. Obviously, if the top-5 signs are no more or less likely to be the correct prediction than any other group of signs, then providing a ranked list is largely useless. In early discussions with potential end--users of this tool, providing a list of ranked suggestions for each sign on a new document was identified as a useful feature, and the top-k accuracy provides a useful measure of success on this task. Given the large dataset imbalance we observe in Section \ref{sec:dataset_summary}, we also wished to characterize classification models' performance as training set attestations vary, to understand which sign classes may be particularly difficult to recognize even in the context of ``perfect'' localizations. For context, deep image classifiers trained on clean, multi-class datasets such as ImageNet achieve top-1 accuracies upwards of 79\%, and top-5 accuracies upwards of 95\% \cite{He_2016} on datasets with far more classes. To understand better how a sign's attestations relate to its per-class precision and recall, we also compute the mean recall, which is commonly used as a "balanced" estimate of accuracy \cite{brodersen2010balanced} and the Spearman's $\rho$ between the number of attestations and a sign's per-class precision and recall. 

\subsubsection{Pipeline}

We evaluate end-to-end transliteration performance in terms of the Character Error Rate (CER), defined as the edit distance between the predicted and true sequences normalized by length. While subtask models may be useful on their own, this metric will determine how well an automated system can produce a whole-document transliteration without human intervention. Modular transliteration systems trained on the 1500-document IAM corpus can achieve CERs under 10 percent  \cite{DBLP:journals/corr/abs-1910-00663}, indicating a high bar for effectiveness on this end-to-end task. We also compute end-to-end metrics analogous to those discussed above - computing the top-k accuracy on signs that could be successfully matched to ground-truth predictions (along with an estimate of the false positive rate of the sign detector).
\section{Methods}

Below we explain the motivation and implementation details behind the design of our computer vision pipeline.

\subsection{Pipeline}

The structure of our modular computer vision pipeline is designed to explore two sets of questions: both to understand performance on the individual subtasks presented by our dataset, as described in section \ref{sec:metrics}, as well as to combine individually trained components to produce an end-to-end OCR system. While it is certainly possible to model this dataset using end-to-end object recognition systems (possibly with an additional linguistic refinement component), the modular structure of this workflow arose from our exploration of individual tasks that could be accomplished within the dataset. 
Our vision pipeline consists of three stages: sign detection, sign classification, and line detection (as shown in Figure \ref{fig:pipeline}). Rather than jointly training a model to produce hotspots and classify signs, we decided to split the problem into two independent modeling steps, inspired by \cite{DBLP:journals/corr/abs-1910-00663}. As such, each stage in the pipeline is trained separately using ground-truth annotations. Potential users of this modeling pipeline identified being able to manually adjust suggested hotspot regions as a key feature, which was facilitated most easily by suggesting hotspots first and then performing inference using a different network. We performed experiments with an end-to-end single-stage RetinaNet (Appendix \ref{sec:singlestage_appdx}), but found similar performance at the cost of modularity and a reduction in interpretability. We leave exploration of additional pipeline variants to future work. We also perform separate analyses of learning tasks within the PFA dataset. These serve to characterize the ``difficulty'' of each learning task (i.e., hotspot localization, sign-classification) given the novelty of this dataset for machine learning training. There may in fact be synergies that an end-to-end system could exploit, such as using contextual information to refine classification prediction, although these could also be achieved using a final, modularized language modeling step. 

\subsection{Sign Detector}
\label{sec:det}

We define cuneiform sign localization as an object detection problem - i.e., selecting a single rectangular bounding box for each cuneiform character. All boxes are assigned to an identical class and an object detection algorithm is trained to distinguish the class against background regions. We use a RetinaNet \cite{DBLP:journals/corr/abs-1708-02002} as implemented in the Detectron2 library \cite{wu2019detectron2} to perform sign localization. The RetinaNet is trained using the Adam optimizer \cite{kingma2015adam}, with an initial learning rate of 1e-4 and a learning rate scheduler that decreased learning rate when performance on validation loss plateaued. Before training, images are rescaled using preset per-channel mean and standard deviation values originally defined on ImageNet. Images are augmented using a mixture of image rescaling, random cropping, and random horizontal flipping and fed into the network in batches of 10. We note that flipping transforms, especially vertical flipping, are problematic in the case of cuneiform character \textit{classification}. Obviously, many cuneiform characters are not invariant to rotations or reflections. However, at the detector stage, the model is not concerned with the specific identity of a character and is designed to recognize any character regardless of value. The network is limited to a maximum of 40000 training iterations, with early stopping performed once the model’s validation performance ceased improving. Final performance metrics on validation folds are computed using the best iteration as judged by AP@50. 

\subsection{Sign Classifier}
\label{sec:cls}

We use a ResNet \cite{He_2016} image classifier to map tablet image regions containing a single sign to a sign class. The classifier maps an image to a 141-dimensional vector of logits, which we use to rank predictions and compute top-k results. Image regions are resized and padded to 50 x 50 pixels before input into the network. During initial experiments, we found that data augmentation (RandomAffine, ColorJitter, and RandomPerspective as implemented in TorchVision \cite{torchvision2016}) as well as standard BatchNorm layers are effective at preventing overfitting, and thus do not apply any kind of weight decay. Early experiments indicated that additional image processing such as Gaussian smoothing and Otsu thresholding \cite{Kurita1992} provided no benefit during training - likely due to such transforms being well-captured by a ResNet's convolutional filters. The classifier network is trained using the Adam optimizer, with initial learning rate 1e-3 for a maximum 500 epochs across all experiments. A learning rate schedule with a patience of 5 epochs is used to reduce learning rate by a factor of 10 when validation loss plateaus. After 10 epochs of no improvement, learning is terminated. The classifier was implemented in PyTorch \cite{NEURIPS2019_9015} using the TIMM library \cite{timm} of predefined model architectures. Statistics were computed at the final epoch of training. Interestingly, we find that using pre-trained models (obviously trained on a very different domain) improve convergence time but not validation performance. To remove unexpected confounding variables, we perform all analyses in this paper on randomly initialized models. We train all classification models on ground-truth, human-annotated hotspot regions and perform tests on both held-out ground-truth hotspots and hotspots predicted by the object detector described in Section \ref{sec:det}.

We experiment with several methods to address severe class imbalance during classifier training, including reweighting per-class loss using inverse frequencies \cite{King01logisticregression}, reweighting with inverse log-frequencies, and using the focal loss \cite{DBLP:journals/corr/abs-1708-02002} that was also used to train the RetinaNet sign detector. However, results are inconclusive (see Appendix \ref{sec:reweight}) and we perform end-to-end inference using a classifier trained using the unweighted cross-entropy loss. 

\subsection{Line Detector}
\label{sec:ransac}

To sort detected hotspots into discrete “lines” facilitating the left-to-right reading of Elamite texts, we use a variant of the Sequential RANSAC algorithm \cite{938625} to iteratively assign elements of the set of hotspot centroids on a tablet to a set of linear models. The algorithm fits a series of L2-regularized RANSAC linear models to the data. Regularization serves to produce lines as flat as possible, and we force each linear model to have a slope less than 0.3 as a further constraint. These parameters, tuned manually, are found to encode our prior belief that lines in this corpus are generally flat or very slightly slanted and rarely intersect. Outliers from the initial RANSAC procedure are used to fit additional RANSAC linear models until there are one or fewer points remaining. Final outliers are assigned to their nearest text line by Euclidean distance. The set of text lines are sorted by their y-intercepts to provide a full left-to-right reading order across multiple horizontal lines. Note that this method of sorting could be problematic for highly slanted lines, which are relatively rare in cuneiform texts, but in practice the majority of the text lines in the PFA are flat enough to avoid significant issues.

\section{Results}

We first perform individual analyses of each stage in the pipeline as trained and tested on ground-truth data, and then perform end-to-end inference and evaluation. We note that the calculated metrics are likely not directly comparable to other work operating on different datasets due to the linguistic differences between them. Our objectives are to assess the utility of the PFA dataset for modern computer vision methods, and we suspect that there are other permutations of computer vision pipelines that can also utilize this data effectively. This particular workflow provides a balance of modularity and interpretability on this dataset, rather than claiming to be a  "universal" tablet image processing pipeline.

We perform by-tablet cross-validation splits, instead of splitting our data randomly by image. Some images contain different regions of the same tablet, and we wish to estimate the performance of our models on entirely unseen tablets, as that most accurately simulates the scenario of performing inference on a novel cuneiform text.  We produce 5 folds, each containing 272 tablets, and one remaining fold containing 271 tablets. Folds contain slightly different numbers of images and hotspots due to differences in text length. The folds are kept consistent across all steps of the modeling pipeline — i.e., the sign classifier and hotspot detector were trained and tested on the same sets of ground-truth hotspots. This ensures that when the sign classifier is used to perform inference on predicted hotspots, the hotspots are unseen by both stages of the modeling pipeline. We then provide metrics aggregated over held-out folds to estimate pipeline performance. We note that when early stopping is performed, held-out fold metrics are used to determine stopping criteria, meaning there may be slight information leakage from the test fold into the fitting procedure. A model evaluation scheme such as nested cross-validation \cite{10.5555/1756006.1859921} would provide more robust error estimates, but we find this to be too computationally expensive for our initial experiments. However, we note that ablation experiments that involve randomly sampling subsets of the data were performed by image, not by tablet. This produces slightly less robust estimates of out-of-sample performance.

\subsection{Sign Detector}

Cross-validation results of the RetinaNet's single-class object detection performance can be found in Table \ref{tab:table_dec}. We observe relatively consistent behavior across folds, indicating that each fold provides sufficient coverage of the sign dataset for the purposes of this training task. We find that increasing the backbone size from 18 ResNet layers to 50 ResNet layers provides no statistically significant improvement in aggregate detection performance, and we also find minor underperformance with 101-layer backbones. This suggests that, on this dataset, it is unlikely that further increases in network depth with the same architecture and loss will provide improvements in performance. As a result,we experiment with using the ResNet18 and ResNet50 backbones for end-to-end experiments in Section \ref{sec:multistage}. 

\begin{table}
	\caption{RetinaNet Detection Performance}
	\centering
	\begin{tabular}{lllll}
		\toprule
		 Backbone & AP@50 &  AP@75 &  Recall@50 \tablefootnote{Here we compute the average of per-tablet recalls at a fixed classification threshold, given the provided IoU threshold.} & Recall@75   \\
		\midrule 
		ResNet18 & 77.4 (2.3) & 20.7 (0.8) & 84.4 (1.6) & 38.3 (1.5) \\
		ResNet50  &  77.1 (1.5) & 20.5 (1.5) & 84.1 (1.7) & 37.7 (1.0) \\
		ResNet101  &  74.4 (3.0) & 12.0 (1.9) & 82.8 (1.6) & 35.5 (1.7) \\
		\bottomrule
	\end{tabular}
	
	\label{tab:table_dec}
\end{table}

\subsubsection{Qualitative Analysis}
\label{sec:det-analysis}
\begin{figure}[h]
\includegraphics[width=0.85\textwidth]{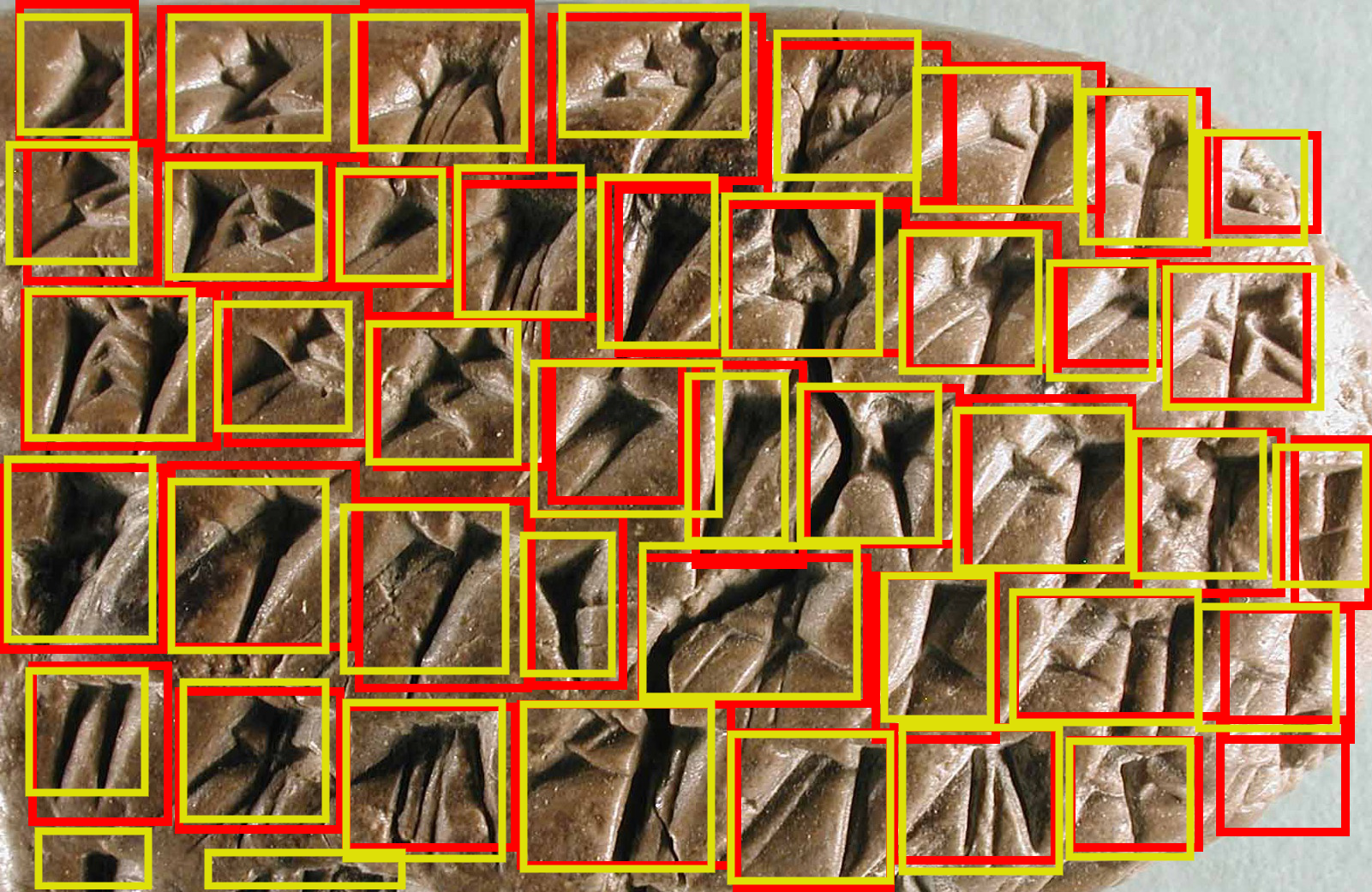}
\caption{Tablet PF0339 - ground-truth hotspots in red, predicted hotspots in yellow.}
\label{fig:qual_example}
\end{figure}

We provide an example of predicted hotspot locations along with their ground-truth counterparts in Figure \ref{fig:qual_example} to illustrate the strengths and weaknesses of our trained sign detection models.


This prediction example displays some of the common pathologies exhibited by the detector—difficulty with oblique signs and false positives (although they appear to be partial signs that are not included in the image), and overlapping boxes. Other common mistakes we observe are the joining of adjacent signs into a compound sign \footnote{Signs that consists of independent subunits but are read as a single sign. These are relatively rare in Elamite, but an example is the compound sign IA which is a concatenation of the signs for I and A. Theoretically, every sign is a compound of the basic wedges “horizontals,” “verticals,” “diagonals,” and an angular hooked wedge known as a “Winkelhaken.” }, and the splitting of compound signs into individual bounding boxes. We hypothesize that integrating linguistic context into the detection stage may reduce these sorts of errors, although that may reduce the modularity of the detection module (i.e., limit it to a language such as Elamite where we have a particularly well-annotated dataset). 

We observe in other images that the detector struggles with signs on edges or sides of tablets, particularly when they are angled away from the camera. This is a consequence of using a 2-dimensional representation of what is ultimately a 3-dimensional object — cuneiform tablets often have signs on their (often curved) sides. We hypothesize that the mis-identification of darkened cracks as signs is due to both the similarity of some thin cracks to simple signs and the appearance of cracked signs in training data.

\subsubsection{Ablation}

\begin{figure}[h]
\centering
\includegraphics[width=0.50\textwidth]{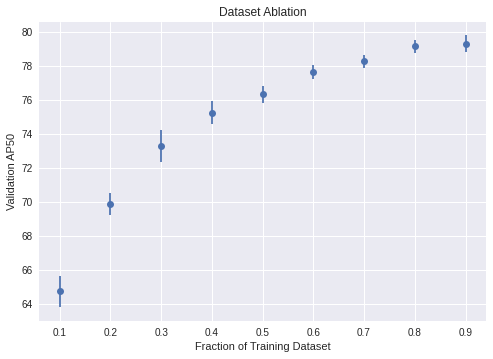}
\caption{Dataset Ablation - Detector}
\label{fig:dec_ablation}
\end{figure}

To investigate the effect of dataset size on detection model performance, we create a single randomized train/test split and randomly subsample the training set at specified fractions. We then evaluate performance on the fixed test split to estimate the marginal utility of adding additional data points to the training set. We note that this experiment was performed before the numeral label adjustments in Section \ref{sec:numerals}, and not re-run due to resource constraints. However, the bounding box dataset is only slightly modified when performing relabeling, so we do not expect significant changes to the trend. 

A plot of detector AP@50 as a function of dataset size can be found in Figure \ref{fig:dec_ablation}. Two patterns are apparent, the first of which is that a detector trained on a small fraction of the PFA dataset performs surprisingly well by this metric. This suggests that lower-quality but reasonable hotspot detectors can be trained using datasets on the order of hundreds of images. It may also imply that a detector can be fine tuned on a related dataset (i.e., another cuneiform tablet set) with relatively few labeled examples, a topic we look forward to exploring in future work. The second is that while AP@50 increases as more data is added, the marginal utility of a new annotated image diminishes as the dataset size increases. \cite{DBLP:journals/corr/SunSSG17} and \cite{joulin2015learning} suggest that there may be a logarithmic relationship between the size of a large computer vision dataset used to train a high-capacity convolutional neural network model and its test performance. While our dataset is certainly not on the scale of current computer vision benchmark datasets, we observe a similar trend.  These results jointly indicate that a RetinaNet object detector is able to produce high-quality segmentations of cuneiform tablets, but that further improvement of the detector is likely bottlenecked by data quality and fundamental limitations of 2D image representations of 3D objects.

\subsection{Sign Classification}

We report sign classification results on ground-truth annotated hotspots in Table \ref{tab:table_cls}, and observe that the top-1, top-3, and top-5 accuracies of ResNet-based classifier models are relatively consistent across folds. We see an increase in classification accuracy as the depth of the ResNet model increases, although the change between an 18-layer ResNet and a 50-layer ResNet is much smaller than that between a 50- and 101-layer ResNet. We also investigate the relationship between a sign's frequency in the training set and its per-class test performance in Table \ref{tab:table_cls_freq}. We find that the mean recall, sometimes referred to as the balanced accuracy \cite{mosley2013balanced}, increases with network depth, although in a diminishing manner. However, we find that the rank-order correlation between the per-class test recall and the class's train frequency is very high. Similarly, per-class test precision is correlated with the class's frequency in training data. This is unsurprising given the extreme imbalance in class frequencies, and provides a limit to the utility of the models—they are likely to be be more useful in automating the annotation of well-attested signs than predicting results for rarer signs. We provide a more detailed exploration of learned sign class representations in Section \ref{sec:classreps}. We also hypothesize that providing sign ranking information could improve the performance of a language model attempting to predict the value of a given sign in context, but we leave this exploration for future work. 

\begin{table}[h]
	\centering
	\begin{tabular}{llll}
		\toprule
		Architecture  &  Top-1 & Top-3 & Top-5  \\
		\midrule 
		
		ResNet18 & 0.657 (0.003) &	0.837	(0.003)	 &	0.888 (0.003)   \\
		ResNet50 & 0.683 (0.006) & 	0.851 (0.006) & 0.897 (0.004) \\
		ResNet101 & 0.692 (0.005) &	0.859 (0.005) & 0.902	(0.004)   \\
		\bottomrule
	\end{tabular}
    \caption{Model Architecture - Classification Accuracy on Held-out Fold}
    \label{tab:table_cls}
\end{table}
\begin{table}[h]
	\centering
	\begin{tabular}{lllll}
		\toprule
		Architecture & Mean Recall & Precision $\rho$ & Recall $\rho$ \\
		\midrule 
		ResNet18 & 0.459  (0.013) & 	0.503 (0.027) & 0.773  (0.040)  \\
		ResNet50 & 0.483 (0.018) & 0.509 (0.049)  & 0.794 (0.042) \\
		ResNet101 & 0.504 (0.022) &	0.519 (0.028) & 0.763 (0.039) 
 \\
		\bottomrule
	\end{tabular}
 	\caption{Model Architecture - Frequency vs Performance. Spearman's $\rho$ values describe the rank-order correlation between a sign's number of attestations and the per-class precision or recall, as a diagnostic tool.}

	\label{tab:table_cls_freq}
\end{table}

\subsubsection{Qualitative Error Analysis}

\begin{figure}[h]
\centering
\begin{minipage}{0.45\textwidth}
\includegraphics[width=0.85\textwidth]{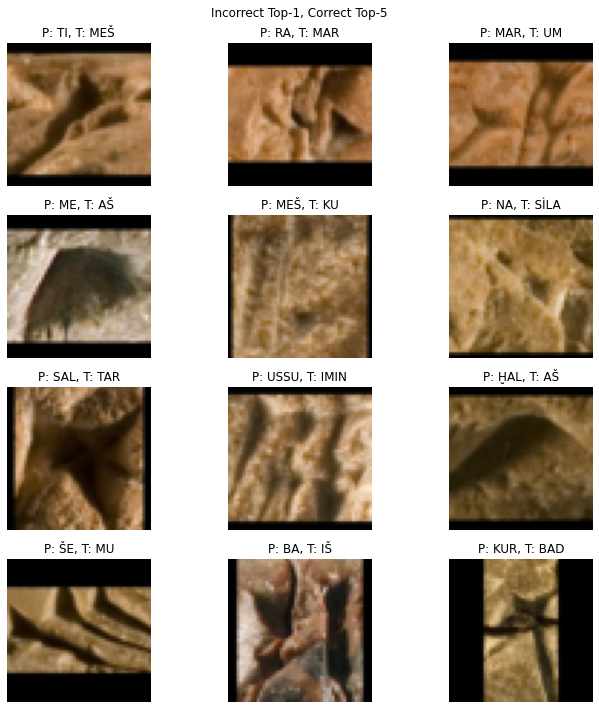}
\caption{Incorrect Top-1, Correct Top-5}
\label{fig:inc_t1}
\end{minipage}
\begin{minipage}{0.45\textwidth}
\centering
\includegraphics[width=0.85\textwidth]{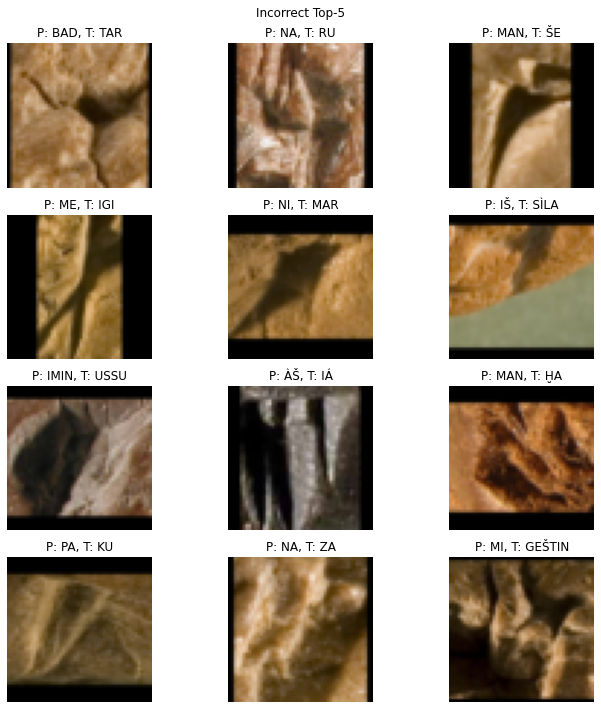}
\caption{Incorrect Top-5}
\label{fig:inc_t5}
\end{minipage}
\end{figure}

Figures \ref{fig:inc_t1} and \ref{fig:inc_t5} contain randomly sampled test-time failure modes of a trained sign classifier on ground-truth annotated hotspots. We point out two types of errors: one where the classifier predicted the top-1 sign incorrectly but the correct sign was within the top-5 predictions, and another where none of the top-5 predicted signs contained the correct sign. A qualitative review of these failure modes indicates that these signs generally would be difficult even for a cuneiformist to identify in isolation. Some signs are nearly impossible to read on photos as they are distorted by the curvature of the writing surface. Others are unrecognizable or hardly recognizable because the sign is damaged, or not all wedges are visible due to poor exposure. Unclear sign borders can lead to misinterpretation, as the sign spacing is highly variable in these texts. The obvious implications of this review are twofold: one, that these classification methods are confused by genuinely difficult cases, and two, that these methods may be useful for automatically flagging ambiguous signs or bad annotations. 

\label{section:cls_perf}
\subsubsection{Per-Class Performance}

We perform a more detailed analysis of the model’s performance on a per-class basis using one test fold as an exemplar. Results for the remaining 4 folds can be found in Appendix \ref{sec:perclass_appdx}. The class distribution of sign images in the PFA dataset is highly imbalanced, implying that performance is be highly variable across classes, and that aggregate top-k accuracy estimates are likely biased by good performance on highly represented signs. This information, in addition to the class rankings discussed above, would be useful to inform users about the expected accuracy of a prediction. We hypothesize that classes that are better represented in the training data are likely to have improved precision and recall, given that high-capacity image classification networks tend to improve in performance (albeit logarithmically) when more training data is provided \cite{DBLP:journals/corr/SunSSG17}. 

\begin{figure}[h]
\includegraphics[width=0.9\textwidth]{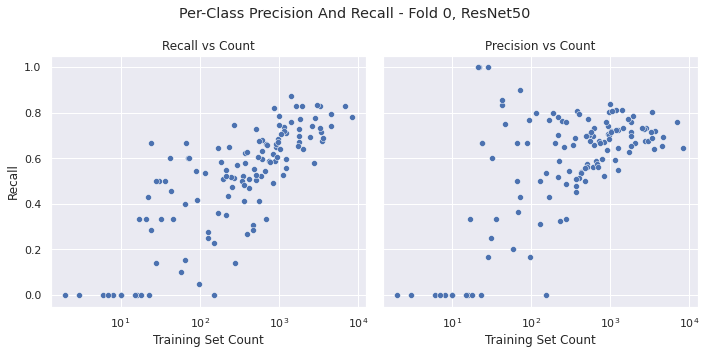}
\caption{Per-Class Precision and Recall}
\label{fig:class_breakdown}
\end{figure}

The relationship between number of training examples and test recall is roughly log-linear (as shown in Figure \ref{fig:class_breakdown}), and a class’ rank in training example counts is highly predictive of its rank in test recall. The log-linear relationship implies that the classifier benefits from additional training examples but the marginal utility of each training example decreases as the training size gets larger. There are also several signs with extremely low, near-zero recall. A few low-frequency signs have very high precisions, and training dataset frequency is less predictive of test precision. Put another way, well-represented signs in the training set are far more likely to be retrieved accurately and completely from the corpus. For example, a prediction of a low-frequency sign is more likely to be a false positive than that of a high-frequency sign. This information could be combined with the top-1 and top-5 classification scores to provide a notion of per-class model uncertainty to an end user. However, the relationship between per-class precision and training set count, which is more useful when evaluating individual predictions, is far less linear than per-class recall.

\subsubsection{Ablation}

\begin{figure}[H]
\centering
\includegraphics[width=0.75\textwidth]{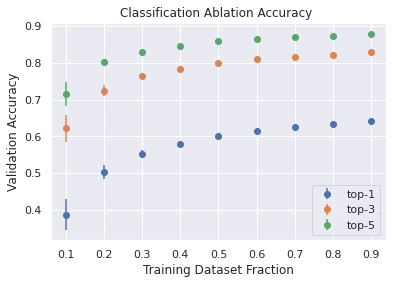}
\caption{Dataset Ablation - Classifier}
\label{fig:cls_abl}
\end{figure}

To investigate the effect of dataset size on classification model performance, we also vary the size of the training dataset while keeping the size of the test data constant. To perform these experiments, we use a single train/test split and randomly subsampled the training data without explicit stratification. We performed 10 subsamples for each sample fraction. Figure \ref{fig:cls_abl} shows test set accuracy as a function of training set sample fraction.

As alluded to in Section \ref{section:cls_perf} above, we see a similar pattern in aggregate accuracy as we do with per-class recall—performance increases as more training examples are added, but the rate of improvement slows down as the dataset gets larger.
We also see that with a very small training sample, the number of classes that are extremely poorly represented (i.e., near-zero recall on test) is high. These results are similar to our observations in Section \ref{section:cls_perf}, indicating a roughly log-linear relationship between sign attestation and predictive performance. These results, in combination with our qualitative exploration of error modes, imply that precisely disambiguating signs is fairly difficult for a classification model due to a combination of label noise, data imbalance, and linguistic ambiguity. Many signs are difficult to interpret for a human with full access to lingustic context, rendering them even more difficult for a model classifying single signs. We see this show up as well in the large gap between top-1 and top-5 accuracies. The model clearly is learning representations that enable meaningful semantic grouping of signs, but we may be hitting a limit of single-sign image classification performance on this type of data. Further dataset refinement to remove unclear signs may aid in this slightly, but we suspect that the models are bottlenecked by image quality and the fundamental difficulty of the task.

\label{sec:classreps}
\subsubsection{Analysis of Learned Sign Representations}

\begin{figure}[H]
\centering
\includegraphics[width=0.75\textwidth]{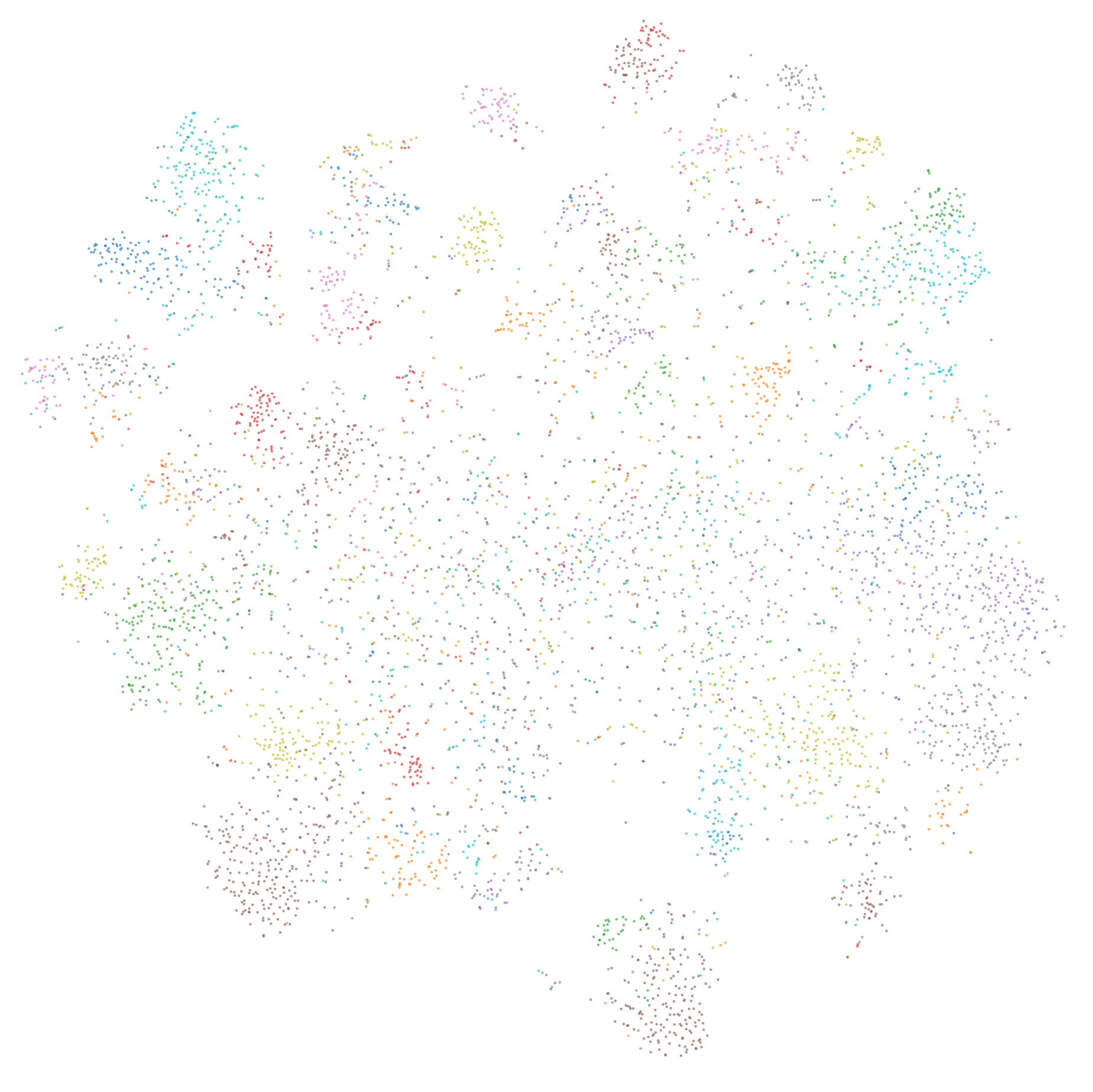}
\caption{Visualization of tSNE Embedding Space.}
\label{fig:tsne}
\end{figure}

Cuneiformists traditionally organize sign lists following an artificially established principle first developed for the Elamite script in 1848. The wedge-shaped elements of signs are analyzed from left to right in a hierarchical fashion—horizontal wedges before diagonals, Winkelhaken, and vertical wedges, and one element before stacks of two, three and so on \cite{borger2004mesopotamisches}. This method is far from ideal as it focuses only on the beginning of the signs and not the overall shape. In contrast, some antique scribes organized their sign lists following the overall shape of the sign (acrographic principle) and typical sign clusters which can appear at the beginning, middle or end of a sign \cite{edzard1982aufbau}. Given that the question of specifically how to organize and group cuneiform signs has great practical relevance for how cuneiform is taught and understood, we want to explore how algorithmically learned representations of signs are clustered.  

To visualize and analyze the sign representations learned by a trained sign classifier, we produce t-SNE plots utilizing a nonlinear dimensionality reduction technique well-suited for visualizing high-dimensional data \cite{JMLR:v9:vandermaaten08a}. Using a trained sign classifier, we embed hotspot images into 141-dimensional logit vectors, encoding the model's unnormalized beliefs about the sign's class.  Then, we perform PCA (Principal Component Analysis) on the full sign classifier dataset (including training data as well as the held-out validation data) to represent the dataset in terms of the top 50 principal components and linearly reduce the data's dimensionality \cite{pearson_pca}. Then we model the dataset's PCA representation with t-SNE to obtain plots that model similar sign instances using nearby points with high probability, and dissimilar sign instances using distant points with high probability. The resulting t-SNE plots show clusters that represent which sign instances the model identifies to be similar. However, it is important to note that cluster sizes and distances between clusters are not necessarily meaningful in t-SNE plots. Individual hotspot images, colored according to their ground-truth class, are embedded using t-SNE and shown in Figure \ref{fig:tsne}  . We perform qualitative analysis of sign clusters to  understand how the learned representations relate, if at all, to the various ways that scholars have grouped cuneiform signs in the past. 

\begin{table}[h]
	\caption{Observed Sign Groups}
	\centering
	\begin{tabular}{lp{5cm}p{8cm}}
		\toprule
		Group & Exemplars \tablefootnote{Paired with Cameron 1948 indexes.} & Comments  \\
		\midrule 
		1 & BAR (21) NU (89) ME (98) MAŠ (99) SAL (102) MEŠ (109) & Dominant vertical at the beginning of the sign in combination with a few other wedges. \\
		\midrule
		2 & UL (90) \tablefootnote{Variant where the first wedge looks vertical. } MI (92) \tablefootnote{Variant starting with a vertical rather than a horizontal.} SAL (102) MEŠ (109) & Close to group one. Dominant vertical followed by groups of horizontals. \\
		\midrule
        3 & MEŠ (109) TUK (112) KU (114) HA (119) EŠŠANA (120) LU  (121) & Connected to group 2 via MEŠ. Otherwise, two or more (HA, EŠANNA, LU)) dominant verticals at the beginning of the sign are grouped with serval horizontals, mostly after the verticals. \\
        \midrule
        4 & IGI (93) KI (95) & Signs begin with a Winkelhaken followed by a vertical. \\
        \midrule
        5 & NUMUN \tablefootnote{This sign is often perceived as a digraph and is sometimes broken across lines as NU + MAN.} (89)  UL (90)\tablefootnote{Variant beginning with a Winkelhaken.} & Connected to group 3 via UL and group 4. Signs begin with a Winkelhaken followed by one or more horizontals. \\
        \midrule
        6 & IR (58) RU (69) ZI2 (86) EŠ5 = 3  IMIN = 7 & Group of three or more (IMIN) dominant verticals. A few IR, RU and ZI2 are included because the grouping of three verticals is dominant in those signs as well. \\
        \midrule
        7 & \breveunder{H}U (30) RI (32) & Both signs have one horizontal followed by two (\breveunder{H}U) or three (RI) verticals and one Winkelhaken at the end.\\
        \midrule
        8 &  RA (15) GI (27) IG (28) PI (62) AM (63) DUB (70) MAR (72) UM (73) GAL (83) TUR (84) &  All signs in this group do have a combination of the sequence of one to three horizontal(s) – one or two vertical(s) – one to three horizontal(s). \\ 
        \midrule
        9 &  NI (57) IR (58) KAL (59) RU (69) UN (75) SA (106) IB (107) and some others & Connected to group 6 via RU and IR. All signs in the group do have a prominent grouping of two staked horizontals and two to three verticals connected with the lower horizontal. Verticals can precede the grouping (SA, IB) or more verticals can be added at the end of the grouping (UN, RU, KAL).\\
        \midrule
        10 & ITI (7) & Closely connected with group 9. ITI is distinct from this group as it has a group of one followed by two horizontals and then three Winkelhaken instead of the verticals. \\
        \midrule
        11 & SU (85) IŠ (52) DA (77) &  Signs begin with two groups of horizontals followed by two to three verticals.\\
		\bottomrule
	\end{tabular}
	\label{tab:tsne_groups}
\end{table}

\begin{table}[h]
	\caption{Example of tSNE Grouped Signs - Cluster 9 }
	\centering
	\begin{tabular}{lllllll}
	\toprule
	NI & IR & KAL & RU & UN & SA & IB \\
	\midrule
	\includegraphics[width=0.1\textwidth]{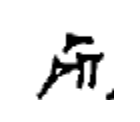} & \includegraphics[width=0.1\textwidth]{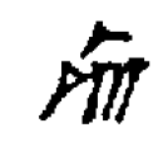}  & \includegraphics[width=0.1\textwidth]{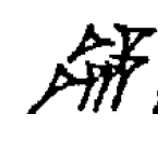} & \includegraphics[width=0.1\textwidth]{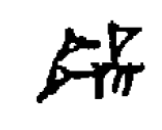} & \includegraphics[width=0.1\textwidth]{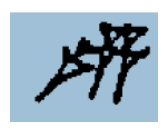} & \includegraphics[width=0.1\textwidth]{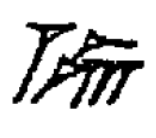} & \includegraphics[width=0.1\textwidth]{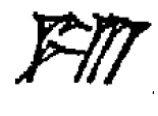}
	
	\end{tabular}
	\label{tab:cluster_9}
\end{table}

Table \ref{tab:tsne_groups} lists the 11 primary sign clusters observed from the 2D t-SNE plots, paired with the predominant signs (labeled using the Elamite sign names devised by \cite{cameron1948persepolis}), along with comments about the grouping. The full t-SNE plot with ground truth sign labels is reproduced in Appendix \ref{sec:tsne_full}. Signs which are adjacent in traditional sign lists do appear together in groupings—for example ME and MAŠ in group 1, IGI and KI in group 4 and \breveunder{H}U and RI in group 7. Groups can also include signs not adjacent or close to each other in traditional sign lists. See for example Table \ref{tab:cluster_9} NI, IR, KAL, RU, UN, SA and IB forming group 9. However, all signs do share prominent grouping of two staked horizontals and two to three verticals connected with the lower horizontal. Despite their departures from traditional sign lists, signs are clearly organized by graphic similarity. Further analysis of learned sign representations could therefore assist in  developing alternate sign lists for didactic purposes. In addition, our preliminary analysis suggests that the models account for variants in handwriting that would be expected from different scribes writing the same signs. The sign UL appears twice, in both group 2 and group 5, based on the starting wedge which can be (correctly) written as vertical or Winkelhaken. We hypothesize that this direction of analysis could be used to identify individual scribes or scribal schools and group tablets by their author.

\label{sec:multistage}
\subsection{End-to-end Pipeline}

Previous sections evaluate classification performance with respect to ground-truth bounding boxes, but ultimately the utility of this computer system will be evaluated by its performance on completely unannotated tablets — i.e., where no ground-truth annotations for intermediate pipeline stages are even present. As such, the estimates of classification performance presented earlier are likely upper bounds. We approach this final evaluation in stages:  first, by evaluating the performance of the sign classifier on bounding boxes predicted by the sign detector, where ground-truth labels have been imputed by alignment to ground-truth hotspots, and finally by evaluating the end-to-end character error rate (CER) of transcribed tablets. Under the existing pipeline, the results provided to the user will consist of predicted bounding box locations for each sign along with a top-5 set of predicted sign identities for each bounding box. Additionally, the pipeline returns a putative horizontal line assignment for each predicted bounding box, which can be used to construct an end-to-end transliteration. This allows users to incorporate the ranked list of sign predictions into their annotation workflow without being overly constrained by the model's top-1 prediction.

Table \ref{tab:table_cls_pred} shows the classification performance of the sign classifier trained on ground-truth bounding boxes, as described in Section \ref{sec:cls}, when performing inference on bounding boxes predicted by the sign detector described in Section \ref{sec:det}. 
We observe a degradation in performance when evaluating on predicted signs, likely due to some of the issues that we observe with the sign detector—erroneous joining or splitting of signs, as well as alignment errors. An end-to-end hotspot detection and labeling training process may provide improved performance here by learning to compensate for shifts in input data distribution induced by the hotspot detection algorithm. To evaluate the CER of transcribed tablets with the current pipeline, we use predicted hotspot locations and top-1 predicted sign labels to predict text ordering using the Sequential RANSAC line detection algorithm described in Section \ref{sec:ransac}. The line detection step groups detected signs into lines, which are ordered top-to-bottom using their Y-intercept. This produces a left-to-right and top-to-bottom reading order sequence of the entire tablet which can be compared to the original tablet's transliteration. We observe a relatively high CER, which indicates that the current pipeline is unable to reconstruct text sequences. A top-1 accuracy of approximately 0.56, along with probable alignment errors, is a promising step towards automatic tablet transcription, although insufficient. 

\begin{table}[h]
	\centering
	\begin{tabular}{llllllll}
		\toprule
		Sign Detector & Classifier & Top-1 Acc.  $\uparrow$ & Top-3 Acc.$\uparrow$ & Top-5 Acc. $\uparrow$ & Detector FPR $\downarrow$ & CER $\downarrow$ \\
		\midrule 
		ResNet18 & ResNet18 & 0.563 (0.010) & 0.735 (0.009) & 0.793 (0.009) & 0.126 (0.011) & 0.683 (0.015) \\
		ResNet18 & ResNet50 & 0.563 (0.010) & 0.735 (0.009) & 0.793 (0.009) & 0.126 (0.011) & 0.684 (0.015) \\
		ResNet50 & ResNet50 & 0.540 (0.011) & 0.710 (0.009) & 0.769 (0.008) & 0.121 (0.012) & 0.686 (0.023) \\
		ResNet50 & ResNet18 & 0.563 (0.007) & 0.734 (0.008) & 0.792 (0.008) & 0.121 (0.012) & 0.669 (0.016) \\
		\bottomrule
	\end{tabular}
 	\caption{End-To-End  performance. Top-K accuracies are computed on predicted bounding boxes that were able to be aligned to ground-truth boxes, with boxes that were not able to be aligned counted as false positives. The sign detector backbone architecture is labeled in the first column. The end-to-end Character Error Rate is also computed. Arrows indicate direction of improved performance.}

	\label{tab:table_cls_pred}
\end{table}

The combination of each model's pathologies, particularly those of the sign classifier, render the combined pipeline unable to accurately transcribe tablets. This result primarily highlights the need for explicit linguistic supervision and the limits of a vision-only approach to tablet transliteration. The noisy outputs of the sign classifier, while useful for providing suggestions to cuneiformists, most likely require effective denoising using linguistic and contextual information. 

\section{Conclusion}

\subsection{Summary}

We have demonstrated that the large and richly annotated Persepolis Fortification Archive enables direct training of computer vision models to recognize and classify Elamite cuneiform signs. A trained object detector can localize cuneiform signs with high precision and qualitatively useful performance. A separately-trained classifier is able to produce high-quality predictions of sign identity. These two systems on their own are sufficient to aid cuneiformists in annotating and identifying signs in Elamite tablet images, although the current iteration of our end-to-end pipeline has insufficient top-1 sign classification accuracy to produce complete transcriptions automatically without any cuneiformist intervention. To our knowledge, this work presents the first computer vision-based analysis pipeline on Elamite cuneiform, demonstrating practically useful performance using a novel dataset.
The pipeline presented in this work will serve as a basis for future work on Elamite cuneiform tablet transliteration. We have released a processed subset of the PFA dataset as well as the source code and trained model parameters for our pipeline. 

\subsection{Future Work}

While our pipeline can effectively localize cuneiform signs and provide a list of suggested readings, the current system was not built to provide complete transliterations. Our future work will seek to rectify this by incorporating linguistic supervision to improve the quality of sign predictions and to provide sign values. This could be achieved in a modular fashion by using a language modeling layer on top of the existing pipeline, or incorporated into the detection phase via a context-aware object detection method such as \cite{Chen_2018_ECCV} and \cite{DBLP:journals/corr/abs-2005-12872}. The PFA in OCHRE contains sign to value mappings for each hotspot annotation, so we anticipate this to be feasible on the dataset. However, the latter approach would likely require adopting an end-to-end sign detection and identification scheme rather than the modular scheme we currently adopt. 

A second avenue of future work consists of applying the hotspot detector to non-Elamite tablets to determine whether or not a hotspot detector ostensibly trained without explicit linguistic information can identify signs in other cuneiform corpora. While it is likely that there is some bias towards the specificities of Elamite cuneiform, preliminary experiments (Appendix \ref{sec:ur3}) have indicated that the detector component of our pipeline can localize signs (without identifying their value) on Ur III Sumerian tablets that predate the tablets in the PFA dataset by over a millennium and a half. However, we suspect that the vast linguistic differences between the corpora will limit the performance of a vision-only or single-language model. The method of \cite{dencker2020deep} was shown to perform reasonably well on a different cuneiform corpus, indicating that there is some transferability between corpora given a trained detector. A combination of our methods and the semi-supervised methods described by \cite{dencker2020deep} may be useful here—for example, using our models (pre-trained on a large annotated dataset) to initialize semi-supervised learning on another cuneiform corpus. 

\begin{acks}
We would like to thank Matthew W. Stolper for his helpful comments, feedback, and for providing our research team with access to the PFA images and transliterations in OCHRE. 
\end{acks}

\bibliographystyle{ACM-Reference-Format}
\bibliography{references}
\appendix
\section{Dataset Statistics}

\subsection{Descriptive Statistics}
\label{sec:appdx_statistics}

\begin{figure}[h]
\centering
\begin{minipage}{0.45\textwidth}
\centering
\includegraphics[width=0.9\textwidth]{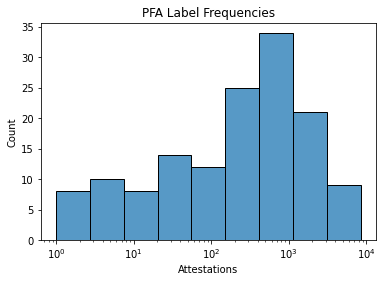}
\caption{Histogram of Sign Frequencies}
\end{minipage}
\begin{minipage}{0.45\textwidth}
\centering
\includegraphics[width=0.9\textwidth]{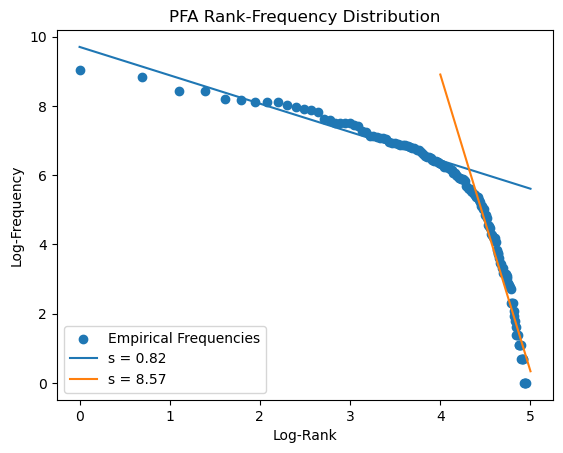}
\caption{Log-Log Rank-Frequency Plot}
\label{fig:label_dist}
\end{minipage}
\end{figure}

The full dataset used in this work consists of 5007 images of 1360 unique tablets, where each image contains a set of hotspot region annotations. Bounded hotspot regions (equivalent to bounding boxes in an object detection framework) are annotated with a categorical label corresponding to the Elamite sign contained within the hotspot. The relative frequencies of the 141 unique Elamite signs contained within the dataset are highly unbalanced, with the vast majority of signs being fairly poorly attested. Power-law behavior of natural language datasets rank-frequency relations has been attested across languages and corpora. Compliance with this behavior, referred to as Zipf's Law \cite{piantadosi2014zipf}, is most simply checked by fitting the log-log rank-frequency data of a corpus to a line (although this method has some error modes, see \cite{doi:10.1137/070710111}). Interestingly, the rank-frequency distribution of the PFA sign data is not well fit by a single power-law ($r^2 = 0.66$ but poor visual fit), but is well described by a broken power law ($r^2 = 0.98$, as shown in Figure \ref{fig:label_dist}), appearing as a piece-wise linear fit in the log-log plots. This behavior is not unexpected from a modern natural language corpus \cite{powers1998applications}. We observe that the PFA corpus exhibits a steep ``fall off'' in the usage of all available signs. The scribes writing Elamite used a script system which developed 2,700 years earlier for Sumerian, an unrelated language. While scribes of the Elamite language employed a common subset of signs, they also had at their disposal a wide range of rare signs, including syllabograms and logograms.

\subsection{Data Preprocessing}
\label{sec:dset}

For further analysis and machine learning experiments, the PFA OCHRE dataset was exported into the JSON file format utilized by the Detectron2 \cite{wu2019detectron2} object detection library. Because tablets were photographed from various angles, not every sign in every image is annotated. To remove large unlabeled regions present in many PFA images, typically consisting of unannotated cuneiform signs, we automatically crop the tablet images using the pixel locations of the tablet's bounding boxes. Hotspot locations are then automatically adjusted to be consistent with these modified dimensions. A subset of the data with low resolution, with signs out of focus, or with signs located on an extreme edge of the tablet, all of which contributed to making the sign illegible even to the human eye, are removed by hand before training. 

\subsection{Data Anomalies}

\subsubsection{Cuneiform Numerals}
\label{sec:numerals}

The PFA project had not identified the cuneiform signs used to record numerals, requiring some intervention in correcting these sign labels. In the cuneiform writing system used by Elamite, numbers are expressed with a a combination of wedges. For example, the number ``11'' is represented by the concatenated signs for ``10'' and ``1". Each of the elements corresponds also to a sign with a syllabic value. 10 is represented by the sign U and 1 is represented by the sign DIŠ.  In the traditional transliterating system also employed by PFA, numerals are transliterated with Arabic numerals, making it sometimes difficult to identify the original cuneiform signs. To avoid mistakes in the recognition of numbers (1 and DIŠ were annotated as separate signs, despite being visually identical), we substitute all simple numerals with their syllabic value 1 = DIŠ and 10 = U. We also remove composite numerals from the dataset, for example 90 (DIŠ + UUU) or 11 (U + DIŠ). This step is performed manually before training is run, removing visually redundant sign classes.

\subsubsection{Broken/Corrupted Signs}

While the majority of the hotspotted PFA tablets are in  good condition, some tablets have suffered damage in the roughly 2500 years in between their creation and their recovery by the OI. Damage in this corpus takes the typical forms observable in other clay tablet corpora: abraded surfaces, prominent cracks, and broken edges or sections of the tablet. In practice, this leads to fairly predictable inference errors—for instance, as discussed in Section \ref{sec:det-analysis}, we observe some false positive detector predictions that consist of misidentifying cracks in a tablet as cuneiform signs.  

\subsection{Per-Fold Counts}

\begin{table}[h]
	\caption{Fold Details}
	\centering
	\begin{tabular}{llll}
		\toprule
		Fold & Tablets & Images & Hotspots \\
		\midrule 
		0 & 272 & 1019 & 19637 \\
		1 & 272 & 1112 & 27819 \\
		2 & 272 & 1039 & 21738 \\
		3 & 272 & 944 & 23272 \\
		4 & 272 & 895 & 23455 \\
		\bottomrule
	\end{tabular}
	\label{tab:folds}
\end{table}

\section{Line Detection Example Output}
\label{sec:ransac_example}

\begin{figure}[H]
\caption{Example Line Detection Output}
\centering
\includegraphics[width=0.50\textwidth]{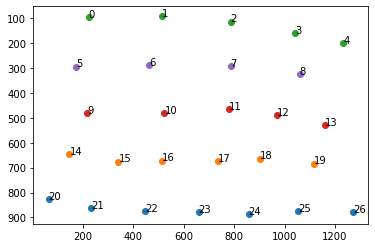}
\caption{Example of Sequential RANSAC line detection performed on ground-truth hotspot centroids from tablet PF 0008. Detected lines are colored, and predicted sign order is displayed next to the centroid.}
\label{fig:line_det}
\end{figure}

\section{Reweighting}
\label{sec:reweight}
We attempted to address this per-class imbalance by performing loss reweighting during model training. We experiment with four loss reweighting methods and evaluate their results on top-k accuracy and per-class precision and recall.

\begin{table}
	\caption{Reweighting - Top-k, ResNet50 }
	\centering
	\begin{tabular}{llll}
		\toprule
		Loss Weighting &  Top-1 Acc. & Top-3 Acc. & Top-5 Acc. \\
		\midrule 
		Unweighted & 0.670	(0.007) &	0.843	(0.006) &	0.892	(0.005)  \\
		Inv. Freq  & 0.368	(0.038) &	0.599 (0.040) & 0.698 (0.0348) \\
		
		Balanced & 0.331 (0.023) & 0.561	(0.025) & 0.663 (0.021) \\
		Log Inv. Freq & 0.663	(0.003) &	0.840 (0.003) & 0.889 (0.003)\\
		Focal & 0.671	(0.005) & 0.845	(0.004) & 0.893 (0.003) \\
		\bottomrule
	\end{tabular}
	\label{tab:table_cls_reweight_topk}
\end{table}

\begin{table}
	\caption{Reweighting - Frequency vs Performance, ResNet50 }
	\centering
	\begin{tabular}{llll}
		\toprule
		Loss & Mean Recall & Precision $\rho$ & Recall $\rho$ \\
		\midrule 
		Unweighted & 0.479 (0.019) & 0.487 (0.073) & 0.767 (0.033) \\
		Inv. Freq & 0.304	(0.034) & 0.922 (0.011) & 0.499 (0.077) \\
		Balanced. & 0.264 (0.020) & 0.921 (0.014) & 0.528 (0.019) \\
		Log Inv. Freq & 0.479	(0.011) &	0.772 (0.036) & 0.609 (0.052) \\
		Focal & 0.472	(0.015) & 0.477	(0.048) & 0.782 (0.050) \\
		\bottomrule
	\end{tabular}
	\label{tab:table_cls_reweight_freq}
\end{table}

We find that simply reweighting according to the frequency of signs in the dataset has a deleterious effect on overall model performance, although it does successfully lower the correlation between training set frequency and test set precision. However, we see a marked increase in the correlation between precision and class frequency. Reweighting according to log-scaled inverse frequencies appears to have little effect on top-k statistics while reducing the correlation between recall and frequency, albeit at the cost of an increase in the correlation between precision and frequency.

\section{Per-Class Precision and Recall Statistics}
\label{sec:perclass_appdx}

\begin{figure}[H]
\caption{Per-Class Breakdown - Fold 1}
\centering
\includegraphics[width=0.5\textwidth]{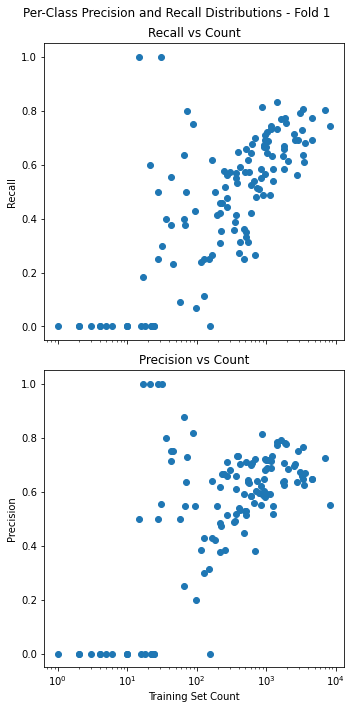}
\end{figure}

\begin{figure}[H]
\caption{Per-Class Breakdown - Fold 2}
\centering
\includegraphics[width=0.5\textwidth]{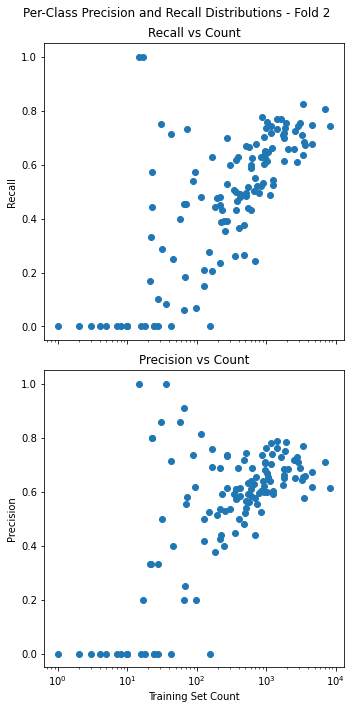}
\end{figure}

\begin{figure}[H]
\caption{Per-Class Breakdown - Fold 3}
\centering
\includegraphics[width=0.5\textwidth]{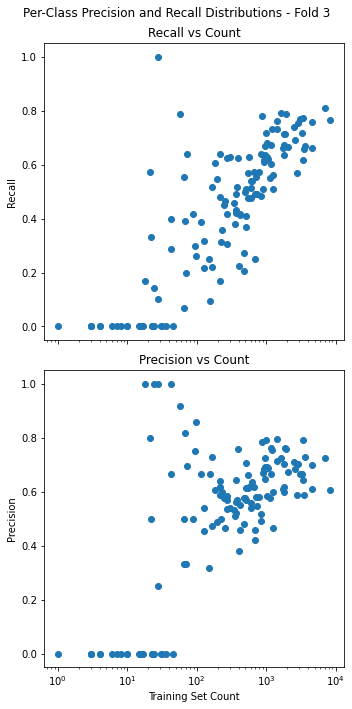}
\end{figure}

\begin{figure}[H]
\caption{Per-Class Breakdown - Fold 4}
\centering
\includegraphics[width=0.5\textwidth]{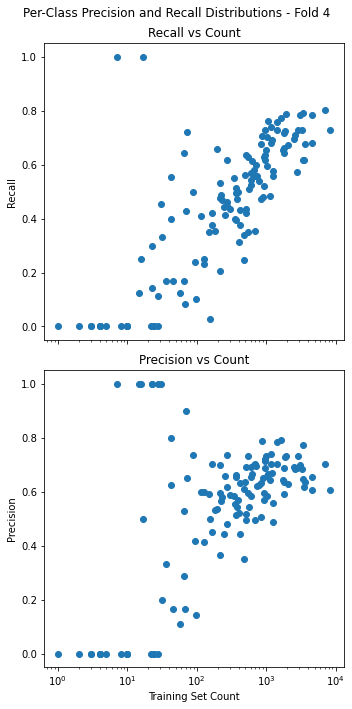}
\end{figure}
\newpage

\section{Full tSNE Plot With Labeled Points}
\label{sec:tsne_full}
\begin{figure}[H]
\caption{Full tSNE Plot With Labeled Points}
\centering
\includegraphics[width=1\textwidth]{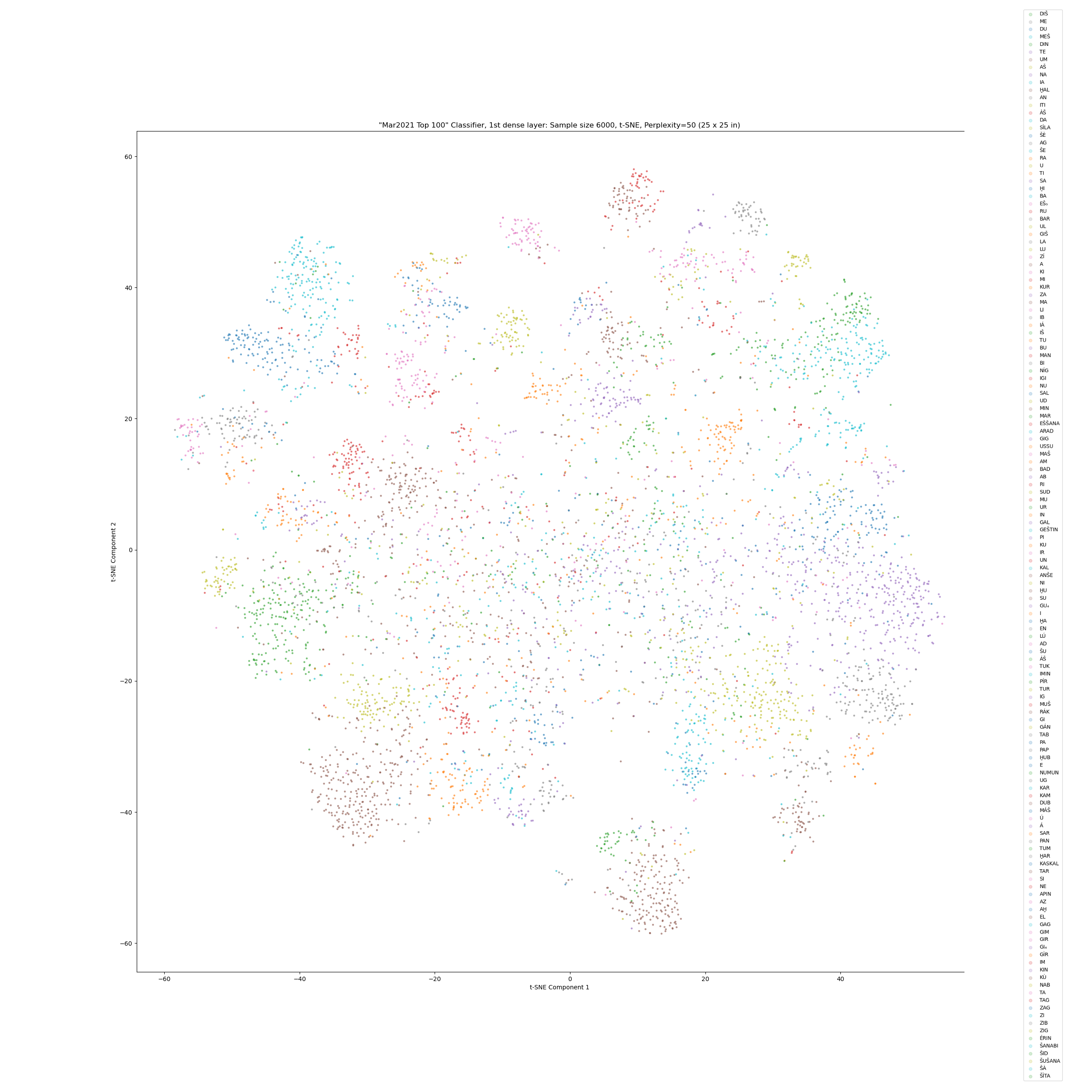}
\label{fig:tsne_full}
\end{figure}

\section{Single-Stage Experiments}
\label{sec:singlestage_appdx}

To compare our multi-stage pipeline to a similar single-stage workflow, we performed an experiment leveraging a multiclass RetinaNet trained on the same corpus. Note that the multi-stage pipeline was trained using the newer, torchvision \cite{torchvision2016} codebase currently available on GitHub rather than the original Detectron2 code. During our initial experiments, we realized that a multiclass RetinaNet represents ambiguity in sign identity by placing many bounding boxes on the same location, as NMS typically does not deduplicate between classes. However, this can be easily remedied by a simple NMS-style algorithm that combines overlapping bounding boxes and uses the confidence scores to produce a set of top-k predictions for each bounding box region.

\begin{figure}[h]
\centering
\begin{minipage}{0.45\textwidth}
\includegraphics[width=0.85\textwidth]{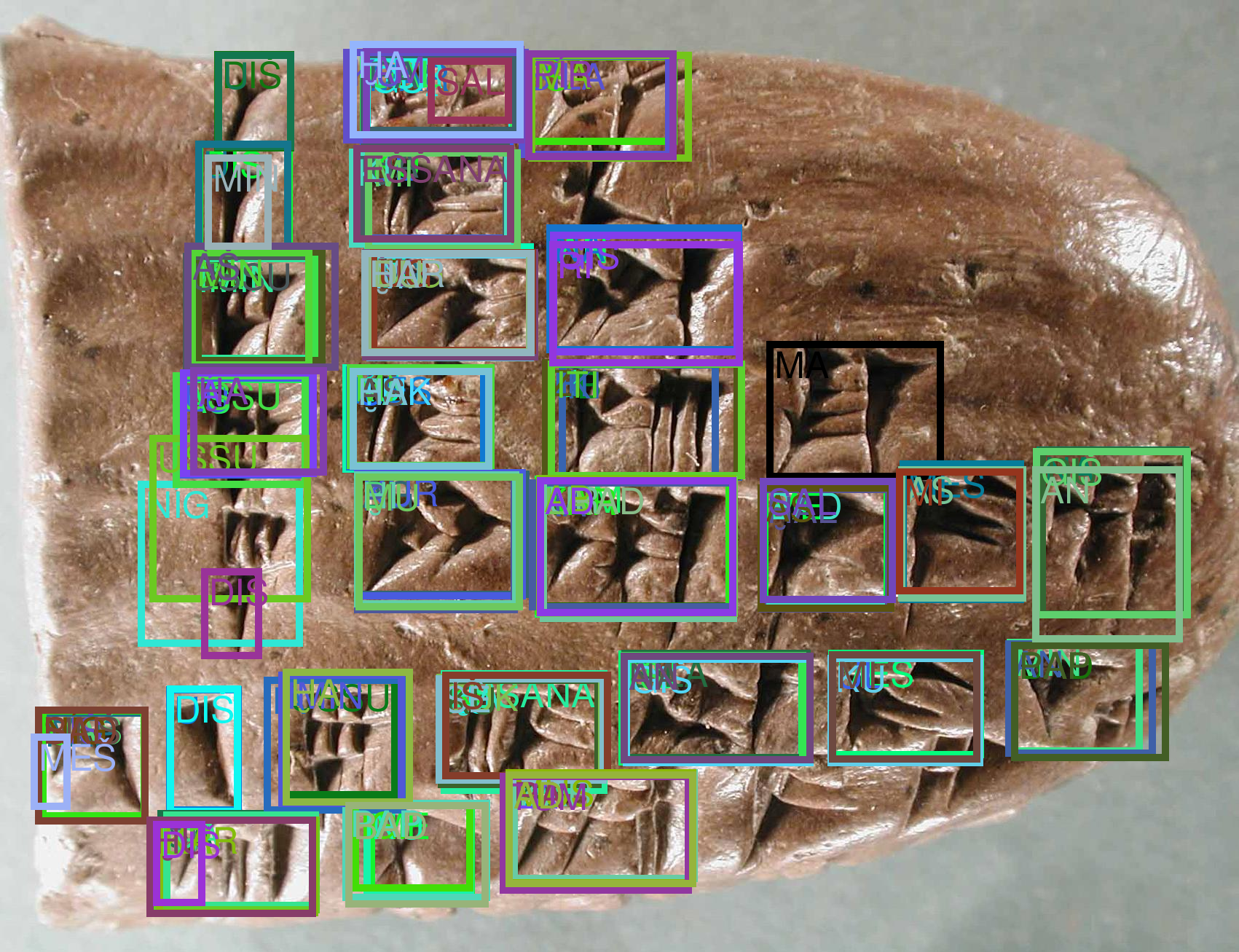}
\caption{Multiclass Detector - Raw Output}
\label{fig:multiclass_raw}
\end{minipage}
\begin{minipage}{0.45\textwidth}
\centering
\includegraphics[width=0.85\textwidth]{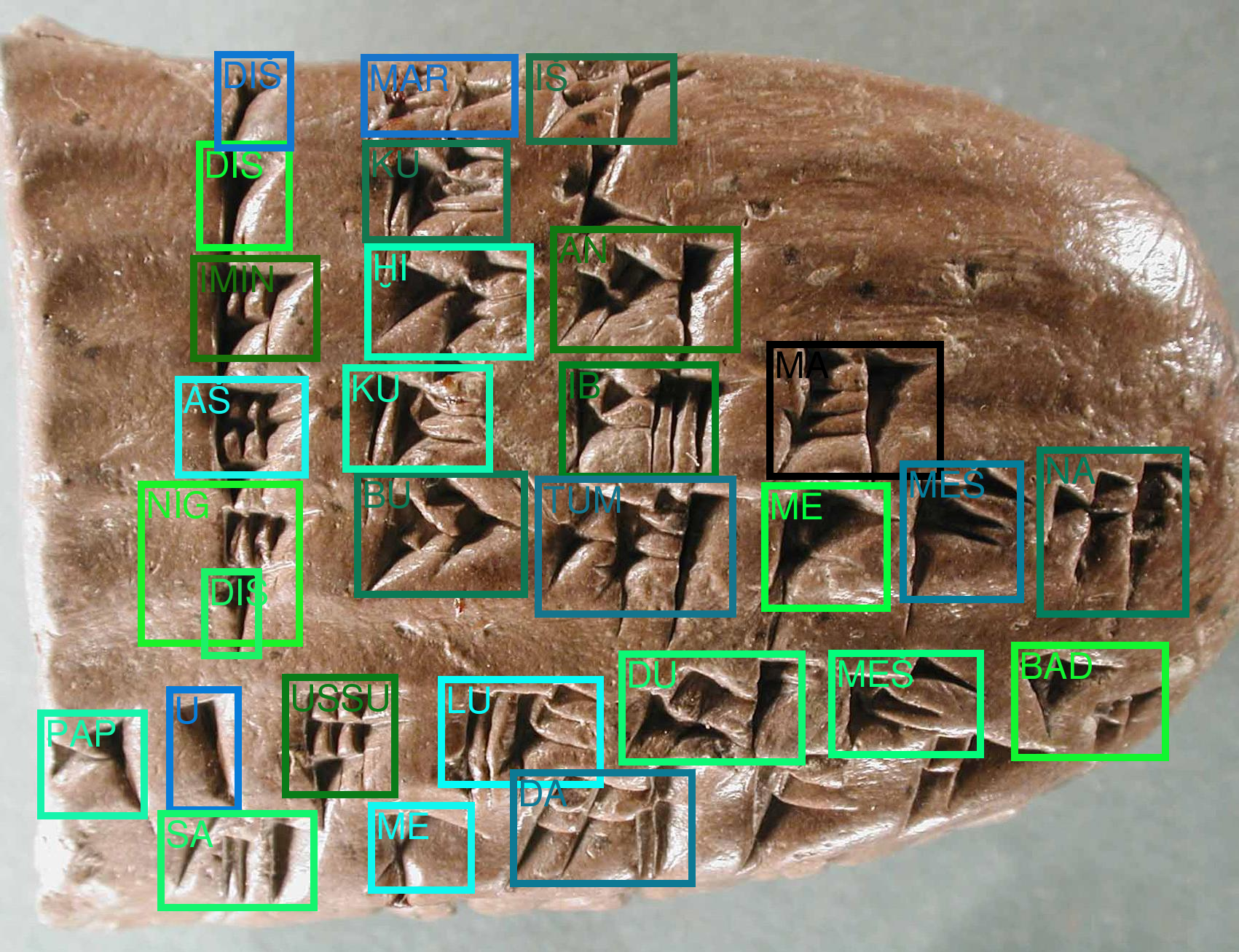}
\caption{Multiclass Detector - Deduplicated}
\label{fig:multiclass_deduped}
\end{minipage}
\end{figure}

We find that aggregate performance is similar to the metrics provided for the multi-stage detector, with Top-1 and top-5 accuracies of 0.58 and 0.73, and a bounding box FPR of 0.14. We hypothesize that fundamental improvements on the end-to-end task will require a different flow of information from image to characters than an object detector can provide. To provide a modular workflow, we focus on the multi-stage pipeline in the body of the paper. 

\section{Generalization to Non-Elamite Tablets}
\label{sec:ur3}

\begin{figure}[H]
\caption{Inference on Ur III Tablet}
\centering
\includegraphics[width=0.75\textwidth]{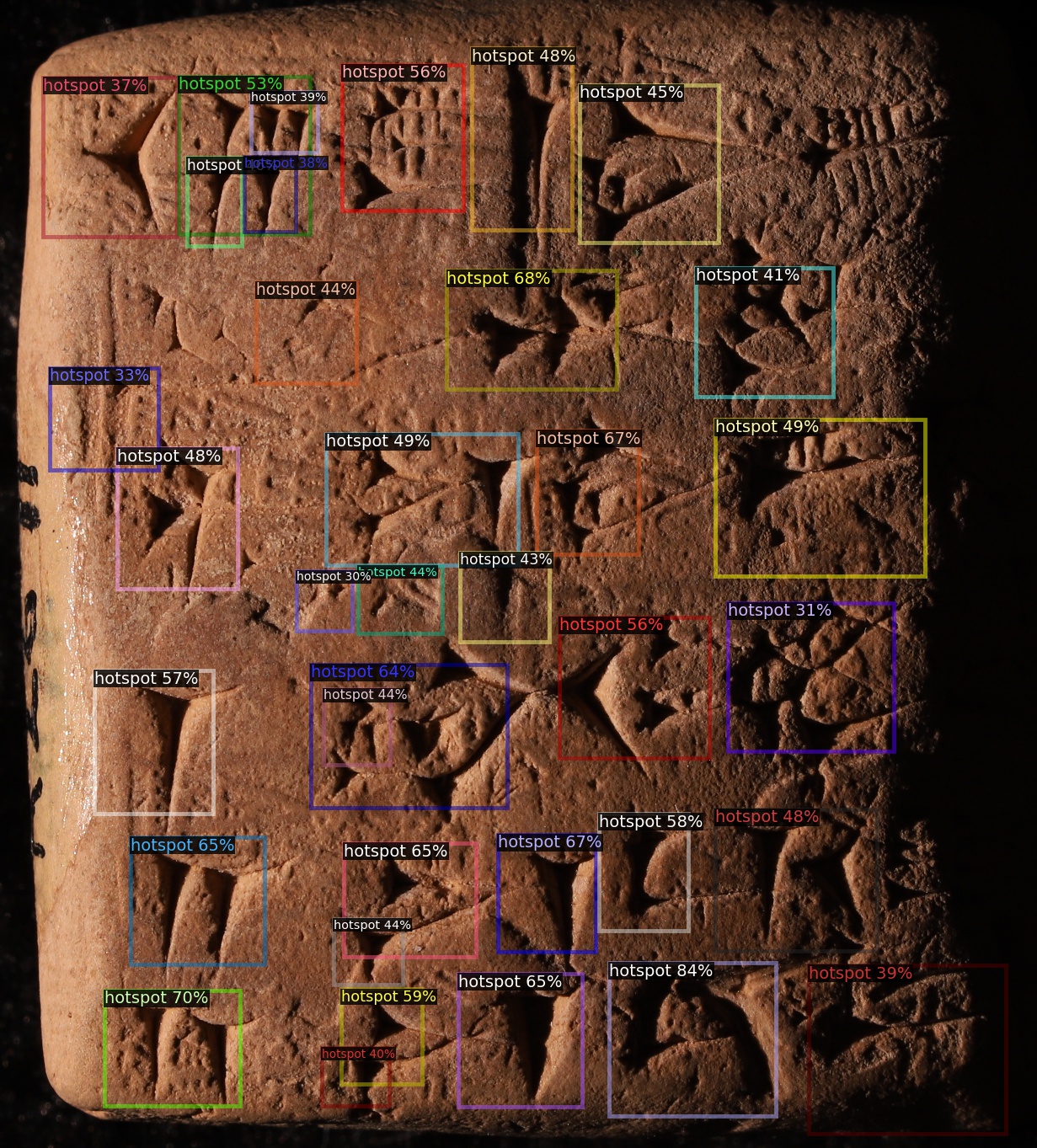}
\label{fig:ur3}
\end{figure}

We hypothesize that a cuneiform sign detector trained on Elamite signs would be able to annotate non-Elamite cuneiform tablets, due to general similarities in cuneiform texts across time periods and languages. The shape, style, and even underlying language changed dramatically over the course of cuneiform's 3000 years as a written language, and we wished to see whether or not the signs were similar enough that an object recognition network trained on one period of cuneiform would be able to recognize another period. This also provides some idea of how useful this pipeline component will be as a general cuneiform sign detector, possibly as an initialization for semi- or weakly- supervised methods such as the method described in \cite{dencker2020deep}. 

Qualitatively, the sign-class-free detector performs fairly well at localizing the cuneiform characters, with issues recognizing signs that are barely visible  or that are dissimilar to Elamite cuneiform (upper right). We believe this is a promising indication that a ``general'' cuneiform sign detector can be produced and adapted to individual script and languages, but further work needs to be done.

\end{document}